\documentclass{article}

\usepackage{acl}

\usepackage{times}
\usepackage{latexsym}

\usepackage[utf8]{inputenc} 
\usepackage[T1]{fontenc}    
\usepackage{hyperref}       
\usepackage{url}            
\usepackage{booktabs}       
\usepackage{amsfonts}       
\usepackage{nicefrac}       
\usepackage{microtype}      
\usepackage{xcolor}         
\usepackage{todonotes}
\usepackage{wrapfig}
\usepackage{cleveref}
\usepackage{caption}
\usepackage{subcaption}
\usepackage{booktabs}
\usepackage{multirow}
\usepackage{longtable}
\usepackage{placeins}

\newcommand*{\affaddr}[1]{#1} 
\newcommand*{\affmark}[1][*]{\textsuperscript{#1}}

\title{mOSCAR: A Large-scale Multilingual\\ and Multimodal Document-level Corpus}

\author{%
  Matthieu Futeral\thanks{Correspondence to \texttt{matthieu.futeral@inria.fr}} \affmark[ \hspace{.2mm} 1,2] \hspace{.23cm}
  Armel Zebaze\affmark[1,4] \hspace{.23cm}
  Pedro Ortiz Suarez\affmark[5] \hspace{.23cm}
  Julien Abadji\affmark[1] \AND
  Rémi Lacroix\affmark[3, 6] \hspace{.23cm}
  Cordelia Schmid\affmark[1,2] \hspace{.23cm}
  Rachel Bawden\affmark[1] \hspace{.23cm}
  Benoît Sagot\affmark[1] \vspace{2.5mm}
  \\
  \affaddr{\affmark[1]Inria} \hspace{.33cm}
  \affaddr{\affmark[2]Département d’informatique de l’ENS, CNRS, PSL Research University}\\
  \affaddr{\affmark[3]Institut du développement et des ressources en informatique scientifique, CNRS} \\
  \affaddr{\affmark[4]Sorbonne Université, Paris, France} \hspace{.33cm}
  \affaddr{\affmark[5]Common Crawl Foundation} \hspace{.33cm}
  \affaddr{\affmark[6]Université Paris-Saclay}\\
}

\begin{document}

\maketitle

\begin{abstract}
  Multimodal Large Language Models (mLLMs) are trained on a large amount of text-image data. While most mLLMs are trained on caption-like data only, \citet{alayrac2022flamingo} showed that additionally training them on interleaved sequences of text and images can lead to the emergence of in-context learning capabilities. However, the dataset they used, M3W, is not public and is only in English. There have been attempts to reproduce their results but the released datasets are English-only. In contrast, current multilingual and multimodal datasets are either composed of caption-like only or medium-scale or fully private data. This limits mLLM research for the 7,000 other languages spoken in the world. We therefore introduce mOSCAR, to the best of our knowledge the first large-scale multilingual and multimodal document corpus crawled from the web. It covers 163 languages, 303M documents, 200B tokens and 1.15B images. We carefully conduct a set of filtering and evaluation steps to make sure mOSCAR is sufficiently safe, diverse and of good quality. We additionally train two types of multilingual model to prove the benefits of mOSCAR: (1)~a model trained on a subset of mOSCAR and captioning data and (2)~a model trained on captioning data only. The model additionally trained on mOSCAR shows a strong boost in few-shot learning performance across various multilingual image-text tasks and benchmarks, confirming previous findings for English-only mLLMs. The dataset is released under the Creative Commons CC BY 4.0 license and can be accessed here.\footnote{\url{https://oscar-project.github.io/documentation/versions/mOSCAR/}}
\end{abstract}

\section{Introduction}

Multimodal Large Language Models (mLLMs) are trained on large amounts of text-image data \citep{radford2021learning,yu2022coca,li2023blip,wang2023cogvlm,OpenAI_GPT4_2023,team2023gemini,team2024chameleon}. The main paradigm until recently was to train a model from a large collection of web-crawled images and their captions \citep{li2021align, wang2022image, chen2022pali}. Models such as Flamingo \citep{alayrac2022flamingo} challenged this paradigm by being additionally trained on interleaved sequences of text and images from web documents, 
showing state-of-the-art results on various tasks and in-context learning capabilities that are not present in models trained on caption-like data only. Additionally, \citet{mckinzie2024mm1} recently proved that including interleaved text-image data during training was necessary to get good few-shot learning performance. However, the datasets used to train mLLMs are either private \citep{alayrac2022flamingo}, monolingual or multilingual but only medium-scale \citep{srinivasan2021wit}. Some attempts have been made to reproduce these datasets \citep{mmc4, laurenccon2023OBELICS} but the resulting datasets are only available in English.

Few image-text datasets are multilingual and most of them are obtained by translating English caption-like datasets, such as multilingual Conceptual Captions \citep{sharma-etal-2018-conceptual}, into multiple languages using neural machine translation (NMT) systems \citep{suris2022globetrotter, maaz2024palo}. This presents some drawbacks such as some languages still being poorly translated by current state-of-the-art NMT models \citep{liu-etal-2020-multilingual-denoising, costa2022no} and some cultural subtleties inherent in each language not being fully conveyed. Some efforts have been conducted to collect large-scale multilingual image captioning datasets, such as LAION-5B \citep{schuhmann2022laion}, but they are limited to caption data too, are relatively noisy and more importantly contain a non-negligible share of ``not safe for work'' (NSFW) content such as p\ae dopornographic images \citep{schuhmann2022laion}.

This motivated us to collect and release the first large-scale multilingual and multimodal document dataset derived from Common Crawl.\footnote{\url{https://commoncrawl.org/}. The Common Crawl Foundation is a non-profit organization that crawls the web on a monthly basis. 
} Our dataset, multimodal OSCAR (mOSCAR), follows the OSCAR initiative \citep{OrtizSuarezSagotRomary2019,AbadjiOrtizSuarezRomaryetal.2021,abadji2022towards} and covers 303M documents in 163 languages, 200B tokens and 1.15B images. Figure~\ref{fig:ex-doc-main} shows an example of a document, more can be found in Appendix~\ref{sec:ex-docs}.
We carry out extensive filtering to increase its safety and quality. To prove mOSCAR's utility, we train a multilingual OpenFlamingo \citep{awadalla2023openflamingo} from a Gemma-2B language model \citep{team2024gemma} on a subset of mOSCAR and captioning data from LAION-400M \citep{schuhmann2021laion}, recaptioned with BLIP \citep{li2022blip}, filtered with CLIP \citep{radford2021learning} and translated with NLLB \citep{costa2022no}. We compare against a similar model trained on captioning data only and show we obtain a strong boost in few-shot learning, confirming previous findings for English \citep{alayrac2022flamingo,mckinzie2024mm1,laurenccon2024matters}. mOSCAR can be accessed here \footnote{\url{https://huggingface.co/datasets/oscar-corpus/mOSCAR}}.

\begin{figure*}[!t]
    \centering
    \includegraphics[width=.9\linewidth]{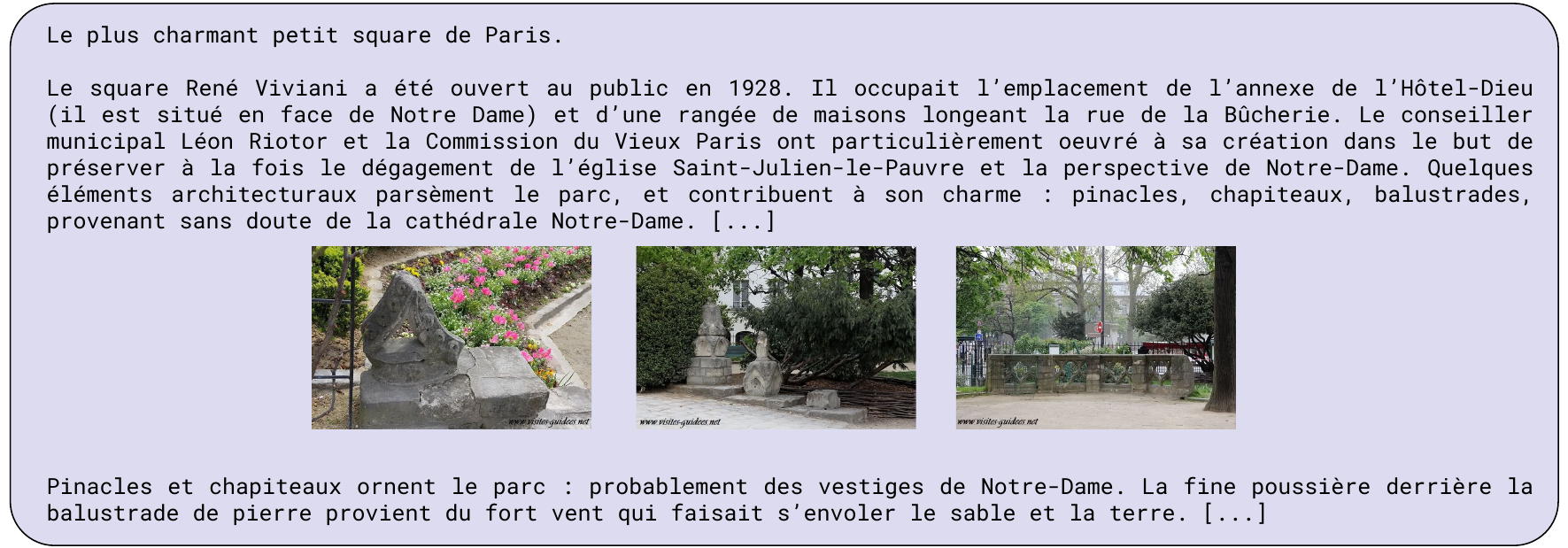}
    \caption{Example of a French document from mOSCAR.}
    \label{fig:ex-doc-main}
\end{figure*}

\section{Related Work}

\paragraph{Large-scale web-based datasets}
Numerous datasets have been created by filtering web-crawled data. These include large-scale text-only datasets \citep{OrtizSuarezSagotRomary2019, raffel2020exploring, wenzek-etal-2020-ccnet, gao2020pile, AbadjiOrtizSuarezRomaryetal.2021,xue-etal-2021-mt5, bigscience_corpus,abadji2022towards,penedo2023refinedweb} and multimodal ones \citep{sharma-etal-2018-conceptual, changpinyo2021conceptual,jia2021scaling,schuhmann2021laion,schuhmann2022laion,kakaobrain2022coyo-700m,laurenccon2023OBELICS,mmc4,gadre2024datacomp}. Even if these datasets are not as high quality as smaller and/or hand-crafted ones, they are now the standard to pretrain foundation models, as it has been shown that training bigger models on more data leads to better downstream performances \citep{brown2020language, hoffmann2022training, touvron2023llama, touvron2023llama2}.

\paragraph{English image-text datasets}
The first open-source image-text datasets were manually created, small-scale and English-only \citep{sbu_captions,lin2014microsoft, plummer2015flickr30k, krishna2017visual}. Scaling up these datasets was an appealing solution to overcome limitations of previous image-text models; a few works \citep{sharma-etal-2018-conceptual,changpinyo2021conceptual} proposed to collect millions of image-text pairs from the web before filtering them with well-designed steps. Relaxing the filtering steps enabled the collection of more data and led to large-scale datasets to train image-text foundation models \citep{radford2021learning,li2021align,schuhmann2021laion, schuhmann2022laion, kakaobrain2022coyo-700m}. However, these datasets generally contain caption-like image-text pairs only, and it is therefore difficult to observe in-context learning abilities similarly to text-only language models trained on raw documents \citep{raffel2020exploring}. \citet{alayrac2022flamingo} overcome this issue by training their model directly on documents with interleaved image-text data. While their results are promising, their M3W dataset is English-only and private. Recently, open-source efforts \citep{mmc4,laurenccon2023OBELICS} have been made to release a similar dataset but they are still monolingual.

\paragraph{Multilingual image-text datasets}
Only a few image-text datasets are available in multiple languages. One of the first focused on collecting Google images from short queries based on word frequencies from Wikipedia pages in 98 languages \citep{hewitt-et-al:2018:Long}. Later, \citet{srinivasan2021wit} proposed the WIT dataset, an image-text dataset composed of Wikipedia pages. Although of high quality, it is only medium-scale even for high-resource languages and there are fewer than 50k unique images for most languages. Another approach lies in bootstrapping multilingual and multimodal data from a model trained with English-only data \citep{pmlr-v189-mohammed23a}. While effective for captioning, it is computationally expensive to implement in practice. Other multilingual image-text datasets exist but focus on captions only and are highly domain-specific \citep{Kosar_2022_BMVC, leong-etal-2022-bloom}.

\section{Dataset Creation Pipeline}\label{sec:data-creation-pipeline}


\subsection{Data collection}

We collect mOSCAR from the Web ARchive Content (WARC) files of three 2023 Common Crawl dumps, processing them using the FastWARC library \citep{bevendorff2021fastwarc}. We remove documents smaller than 500 bytes (50\% of the documents), as we find they are usually too small to be considered documents and tend to contain noisy text. 
We then navigate through the entire Document Object Model (DOM) tree with a depth first search algorithm and ChatNoir library \citep{bevendorff:2018} to extract nodes of interests corresponding to specific HTML tags. 

Following previous work, we extract text from the tags that usually contain the main content of web pages (we refer to them as DOM text nodes), i.e.~
\texttt{<p>}, \texttt{<h*>}, \texttt{<title>}, \texttt{<description>}, \texttt{<ul>}, \texttt{<ol>}, \texttt{<aside>}, \texttt{<dl>}, \texttt{<dd>}, \texttt{<dt>}. 
Similarly to \citep{laurenccon2023OBELICS}, we choose to remove \texttt{<table>} content as most often it is irrelevant and difficult to render. We extract all \texttt{<img>} tags (we refer to them as DOM image nodes). We then remove documents with fewer than 3 text nodes (as they do not contain enough text) and more than 30 image nodes (as we found them to be too noisy). 

\subsection{Language identification}\label{ss:language-id} We identify the language of each document using the state-of-the-art open-LID language detector \citep{openlid}, covering 201 languages. We apply open-LID to each DOM text node and keep the three most probable languages with their respective probabilities. The language of the document is then determined by summing over the probabilities of each language detected for each text segment, weighted by 
the number of characters in the segment\footnote{This is to avoid mis-assigning the language due to the presence of many short, non-informative DOM text nodes in the same language (e.g.~``Cookies'', ``Subscribe'', ``Newsletter'' etc.) and because language identification is generally less reliable for short segments.} 
and taking the language with the highest score.

\subsection{Text-only filtering}

We apply a series of filtering steps to the text content of each document independently of the images, with the aim of discarding poor quality documents and cleaning text as best as possible. We first filter at the text-node level and then at the whole document level, before running near-deduplication to keep unique text nodes within a document and unique documents in the dataset.

\paragraph{Text node filtering} 
We use a set of heuristics (see Appendix~\ref{sec:heuristics}) to extract as much human-generated content as possible while discarding noisy text related to ads and website functions (e.g.~``Instagram'', ``Facebook''). We then keep DOM text nodes with content over 10 bytes. This step, designed to improve the quality of extracted text, removes on average 55\% of text nodes.



\paragraph{Document filtering}
We mostly filter ``not safe for work'' (NSFW) content at the document level. We use an English regular expression to detect adult content, similar to the one used by the Université Toulouse 1 Capitole\footnote{\url{https://dsi.ut-capitole.fr/blacklists/index_en.php}} and remove the entire document if there is a match with any of the DOM text nodes' contents, removing on average 0.5\% of documents (mostly English ones). We acknowledge that there is a high probability that this also discards safe content, e.g.~we could remove content from certain communities who use some explicit words in a non-sexual way \citep{sap-etal-2019-risk}. However, we explicitly favour recall over precision to minimise the risk of unsafe content. 
We additionally remove documents containing fewer than five DOM text nodes and fewer than 300 characters after the previous filtering steps, removing 70.6\% of documents.

\paragraph{Deduplication}
We conduct several types of per-language deduplication at different levels, as this has been shown to improve training efficiency \citep{abbas2023semdedup}. First, we keep unique documents only by removing exact duplicates at the document level. We also remove exact duplicates of text nodes within the same document (4\% of text nodes) and near-duplicate text nodes (1\% of text nodes) by computing the Levenshtein ratio \citep{Levenshtein_SPD66} between all text nodes within the same document and applying a threshold of 0.95. If near-duplicates are found, we keep the first one in the document. 
Finally, we conduct per language near-deduplication at the document level with MinHashLSH \citep{minhash, lsh} following \citet{smith2022using}, removing on average 19\% of documents:\footnote{With some disparity among languages as we found more duplicates for low- than high-resource languages.} we turn documents into hashing vectors, compute min hashes from these vectors and perform Locality Sensitive Hashing to remove duplicates\footnote{We performed this using the \texttt{datasketch} python library.} (see Appendix~\ref{sec:text-dedup-lsh} for more details). 

\paragraph{Toxicity filtering}

Toxic content targeting individuals or groups of people is widespread on the internet and can therefore be found in large-scale web-crawled datasets like mOSCAR without appropriate filtering steps. To alleviate this issue, we apply the same method used by \citet{costa2022no} and remove documents from mOSCAR based on the presence of a list of ``toxic'' words for each language\footnote{The list of these words for each language can be found here: \url{https://github.com/facebookresearch/flores/tree/main/toxicity}}. As some words in the list can also be used in a non-toxic way based on the context (e.g.: `breast' in English), we tag the document as toxic and remove it from mOSCAR if it contains at least two distinct words in the list. This filtering step removes 0.95\% of the documents for very high-resource languages (>5M documents), 2.13\% for high-resource languages (<5M, >500K), 0.47\% for mid-resource languages (<500K, >50K) and 0.64\% for low-resource languages (< 50K). When manually analysing 1,000 random documents removed by this filtering step in each of the 2 (high-resource) languages we are native speakers of (English and French), we found 568 documents with toxic content.

\paragraph{Personal Identifiable Information} 

Personal Identifiable Information (PII) can be found in large-scale web-crawled datasets, we therefore conducted an additional filtering step to replace all detected PII by place holder strings using regular expressions (Appendix~\ref{ss:pii} for more details). Concretely, we replaced all detected email addresses, phone numbers, credit card numbers, IP addresses and passport numbers. Moreover, it was shown that CommonCrawl contains a non neglictible part of API keys in its content \footnote{\href{https://trufflesecurity.com/blog/research-finds-12-000-live-api-keys-and-passwords-in-deepseek-s-training-data}{https://trufflesecurity.com/blog/[...]}}. We therefore scanned the dataset with the trufflehog tool \footnote{\url{https://github.com/trufflesecurity/trufflehog}} to track down residual API keys that could have passed previous filters. We found $\sim$200K positive matches and manually check a random sample of 1K positive matches. We found only 2 of them to potentially be API keys, other matches are mainly noisy text nodes not related to PII. We removed the 200K text nodes from mOSCAR.

\subsection{Image-only filtering} \label{sec:img-only-filtering}

We downloaded images from the URLs in DOM image nodes 
using a modified version of the img2dataset toolkit \citep{beaumont-2021-img2dataset} that includes an antivirus scan and follows \texttt{robots.txt} instructions to respect the Robots Exclusion Protocol. We then apply a series of filtering steps, first removing images based on heuristics, and then applying multiple NSFW detection models to remove undesirable content. Finally, we conduct a set of deduplication steps.

\paragraph{Rule-based filters}
Similarly to previous works \citep{schuhmann2021laion} and to avoid extracting low-resolution images and favicons, we keep images with a minimum height and width of 150 pixels. We restrict the aspect ratio to be between 3 and 1/3 (to remove banners), we remove images if their URLs contain the words ``logo'', ``banner'', ``button'', ``widget'', ``icon'' or ``plugin'' or if the image name from the URL matches ``twitter'', ``facebook'' or ``rss'' (to remove logos). This step removes 13.6\% of the URLs.
At this stage, we downloaded 2.5B images with an average success rate of 55\%.

\paragraph{NSFW detection} We use multiple NSFW automatic models to remove as much unsafe content as possible. We first combine two NSFW detectors: nsfw-detector \citep{nsfw}, a 5-class classifier with a MobileNet \citep{howard2017mobilenets} backbone fine-tuned on 60GB of annotated data and NudeNet,\footnote{\url{https://github.com/vladmandic/nudenet}} an object detector trained to detect different types of nudity in images. We combined the two models as we found the first to be gender-biased while the second gives a large number of false positives for non-human images. Concretely, we consider an image an NSFW candidate if the sum of the probabilities for the classes `porn' and `hentai' is superior to 0.8 using nsfw-detector. We then tag the image as NSFW if one of the sensitive `exposed' classes of NudeNet gets a probability superior to 0.5. 

If a document contains an image with an NSFW tag, we remove the entire document from the dataset, which removes 0.5\% of images. We manually inspecting 1,000 images of the remaining data and found no NSFW content. We manually inspected 1,000 images of the removed content and found 63.4\% of NSFW images.

\paragraph{CSAM content} Child Sexual Abuse Material (CSAM) is widespread on the internet and is therefore likely to be found in such a large-scale dataset crawled from the web. Removing CSAM is challenging as there is no training data nor open-source detection models available as these could be used in a harmful way. We rely on Thorn's CSAM classifier\footnote{\url{https://safer.io/}}, a proprietary classifier trained to detect CSAM content in images. We tag the image as CSAM if the probability of the class CSAM is superior to 0.4 to favour recall over precision. As mentioned above, if a document contains an image with a CSAM tag, we remove it from the dataset. This step removes 0.07\% of the images.

\paragraph{Deduplication} To avoid memorisation issues often seen in models trained on datasets with many duplicated images \citep{somepalli2023diffusion, carlini2023extracting, webster2023duplication, somepalli2024understanding}, we perform deduplication at the image level. We first remove duplicate images within the same document by URL matching (removing 8.7\% of URLs). We then compute a perceptual hash (pHash) for each image using the imagehash library\footnote{\url{https://github.com/JohannesBuchner/imagehash}} and remove images with the same pHash within the same document, keeping only the first occurrence. We also limit the number of times an image can appear in the dataset per-language to 10 using both URL matching and perceptual hashing (this removes 2.5\% of images). We do this per-language and not across languages as having the same images in documents from different languages could encourage cross-lingual transfer. 

\paragraph{Personal Identification Information}

To protect PII in images, we use a lightweight face detector\footnote{\url{https://github.com/Linzaer/Ultra-Light-Fast-Generic-Face-Detector-1MB}} and apply a threshold of 0.99 to detect faces in the images. We apply such a high threshold as we found the model to be biased towards detecting faces with high probability in images without any human. For each image in mOSCAR, we distribute the bounding boxes of the detected faces so that users can blur them when downloading the images. More details are provided in Appendix~\ref{ss:pii}.

\subsection{Data decontamination}

LLMs and mLLMs are trained on web-crawled data that can contain the benchmarks they are tested on \citep{dodge-etal-2021-documenting}. As they are good at memorizing training data \citep{carlini2023extracting}, this data contamination is problematic. 
We therefore discard all images with the same perceptual hash as any of the images from the evaluation benchmarks (and their training sets) we use (see Section~\ref{sec:eval}). This step removes on average 126,016 images for high-resource languages (up to 300K images for English), 6,862 images for mid-resource languages and 45 images for low-resource languages. 

\subsection{Text-image joint filtering} \label{sec:txt-img-joint-filtering}

Our aim is to obtain truly multimodal documents where all images are related to at least one of the text nodes in some way\footnote{We do not limit ourselves to caption-like relation and instead allow all types of text-image relation.} and vice versa.
We choose to apply joint text-image filtering to discard images and/or text nodes that are irrelevant to the rest of the document (e.g.~the case of ads and website functionalities).  

To do this, we use NLLB-SIGLIP\footnote{\texttt{siglip-base-patch16-224} as vision encoder and \texttt{nllb-distilled-600M} as text encoder.} \citep{visheratin2023nllb}, a multilingual version of SIGLIP \citep{siglip} trained with the encoder of NLLB \citep{costa2022no}, which covers all mOSCAR languages.\footnote{We use the open-clip \citep{ilharco_gabriel_2021_5143773} model version and the transformers \citep{wolf-etal-2020-transformers} library.} 
We compute cosine similarity scores between all images and all paragraphs\footnote{We refer to paragraph as the text content in a DOM text node.} within a same document. To remove irrelevant text nodes or images in a document, we mimic a text-image retrieval task, which means we avoid using arbitrary cosine similarity thresholds for each language and can reduce length biases and those in favour of caption-like paragraphs.
For each candidate pair we randomly sample 63 negative images and 63 negative similar-length paragraphs from the same language but other documents. We tag the text node (resp. image) as valid if the cosine similarity of the pair is among the top 8 of the text-to-image (resp.~image-to-text) similarity scores computed with the candidate text node (resp. image) and all the negative images (resp.~text nodes). This means that we tag the text node (resp.~image) as valid if it has a significantly higher score than a score computed with a random image (resp.~text) for at least one of the images (resp.~text node) in the document. We then discard text nodes and images not tagged as valid (on average 35\% of the DOM text nodes and 10\% of the images within a document). 

After this filtering step, we apply additional text-only filters to keep documents superior to 100 bytes. We also reapply the open-lid language detector \citep{openlid} as described in Section~\ref{ss:language-id} as we found the last filtering step to change the major language of some documents.

\begin{figure*}[htbp]
    \centering
    \vspace*{-5mm}
    \begin{subfigure}{0.32\textwidth}
        \centering
        \includegraphics[width=\textwidth]{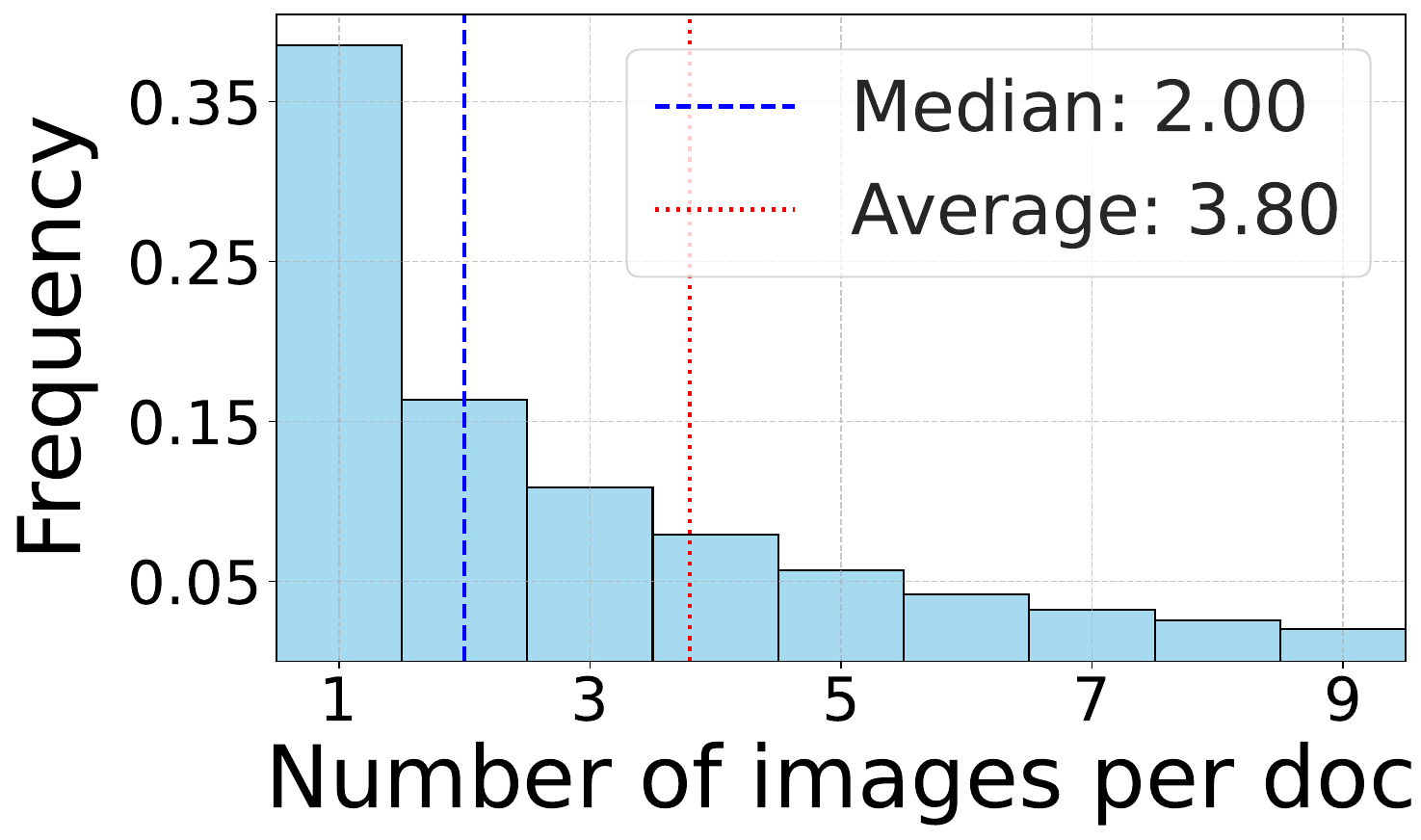}
        \caption{Number of images}
        \label{fig:distrib-img}
    \end{subfigure}
    \hfill
    \begin{subfigure}{0.32\textwidth}
        \centering
        \includegraphics[width=\textwidth]{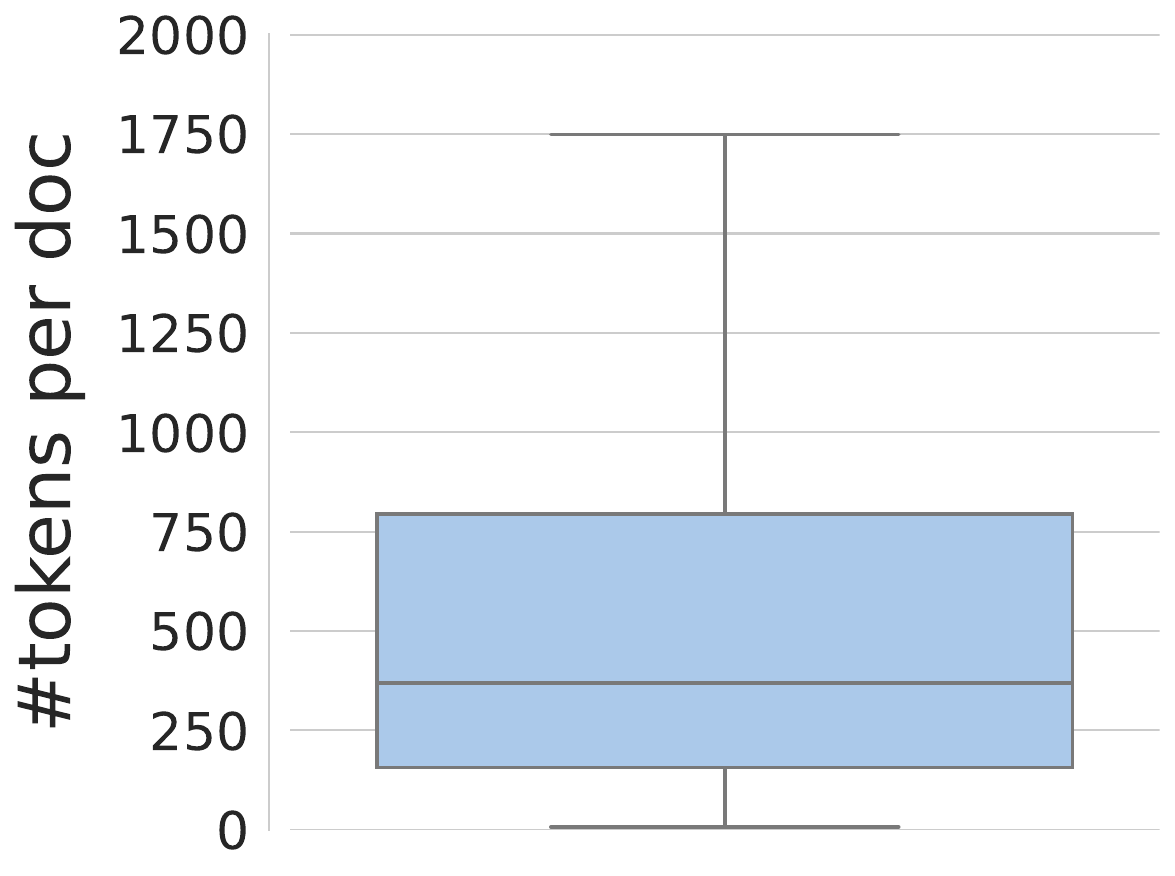}
        \caption{Number of tokens}
        \label{fig:distrib-tokens}
    \end{subfigure}
    \hfill
    \begin{subfigure}{0.32\textwidth}
        \centering
        \includegraphics[width=\textwidth]{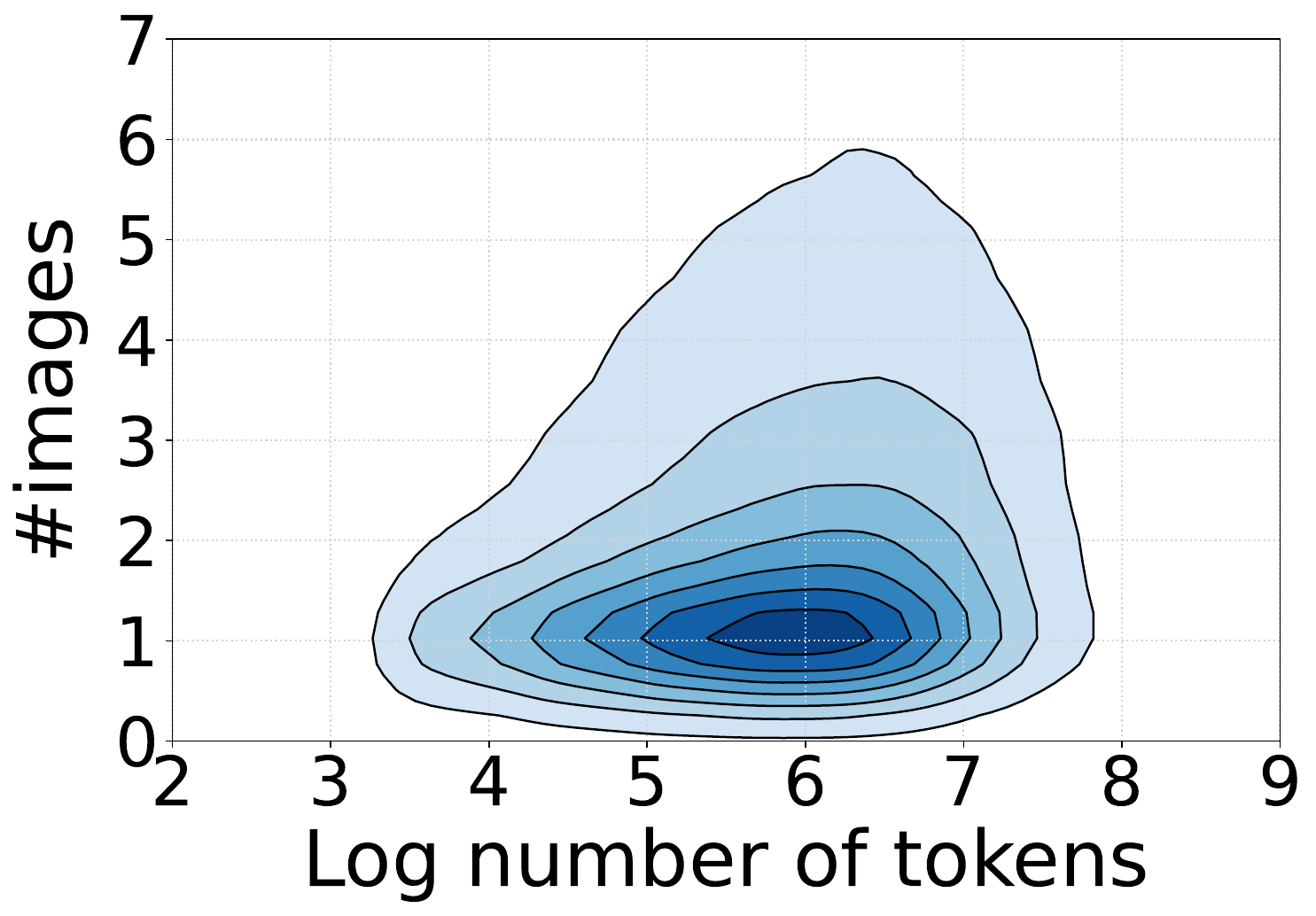}
        \caption{Number of images and tokens}
        \label{fig:joint-distrib-img-tokens}
    \end{subfigure}
    \caption{Distributions of numbers of tokens and images per document.}
    \label{fig:global-stats}
\end{figure*}

\section{Multimodal Open Super-large Crawled Aggregated coRpus (mOSCAR)}

mOSCAR is extracted from three Common Crawl dumps from 2023. Due to computational constraints and in order to extract a maximum number of documents for low-resource languages, we extracted all languages from the first dump only. We removed the 6 most high-resource languages from the second dump and only extracted the languages with fewer than 1M documents for the last dump. Table~\ref{tab:num-docs} shows a distribution of the total number of languages and their number of documents.
To avoid data poisoning \citep{carlini2023poisoning}, we release a hash (sha512) with each mOSCAR image. mOSCAR is composed of 303M documents (200B tokens, 1.15B images) from 163 languages. 
Figure~\ref{fig:global-stats} shows the distribution of images and tokens per document and their joint distribution. 
As shown in Figure~\ref{fig:distrib-img}, the mean and median number of images per document is 2 and 3.80. 

\begin{table}[!ht]
    \centering\small
    \resizebox{.47\textwidth}{!}{\begin{tabular}{lccccccccc} \toprule
     \#docs.  &  10M  & 5M & 1M & 500K & 200K & 50K & 10K & 5K & 1K  \\ \midrule
    \#langs. &  10  & 15 & 36 & 46 & 56 & 75 & 119 & 133 & 154 \\ \bottomrule
    \end{tabular}}
    \caption{Number of languages with at least $N$ documents}
    \label{tab:num-docs}
\end{table}



\subsection{Quality vs Diversity}

While improving overall data quality, the filtering steps we applied (see Section~\ref{sec:data-creation-pipeline}) necessarily have a negative impact on diversity. We therefore study the trade-off between quality and diversity and compare against previously published, well-used datasets. 

\subsubsection{Text content}
\paragraph{Diversity} By contruction, mOSCAR is diverse in terms of number of languages, so we focus on the diversity of mOSCAR's English documents and compare against mmc4 \citep{mmc4}, OBELICS \citep{laurenccon2023OBELICS} and the English subset of WIT \citep{srinivasan2021wit}. We compute the Vendi score \citep{friedman2023the} on a set of SimCSE embeddings \citep{gao-etal-2021-simcse} with a RoBERTa encoder \citep{liu2019roberta} to evaluate the content diversity.  
Since embedding-based diversity metrics target content diversity well but are less relevant for lexical diversity \citep{tevet-berant-2021-evaluating}, we measure lexical diversity via the distinct $n$-gram ratio \citep{li-etal-2016-diversity}. An analysis of the topics \citep{grootendorst2022bertopic} found in multiple languages of mOSCAR where we show diverse topics across languages can be found in Appendix~\ref{sec:topic-modeling}.

\paragraph{Comparison with other datasets}

\begin{figure}[!h]
  \centering
  \includegraphics[width=.4\textwidth]{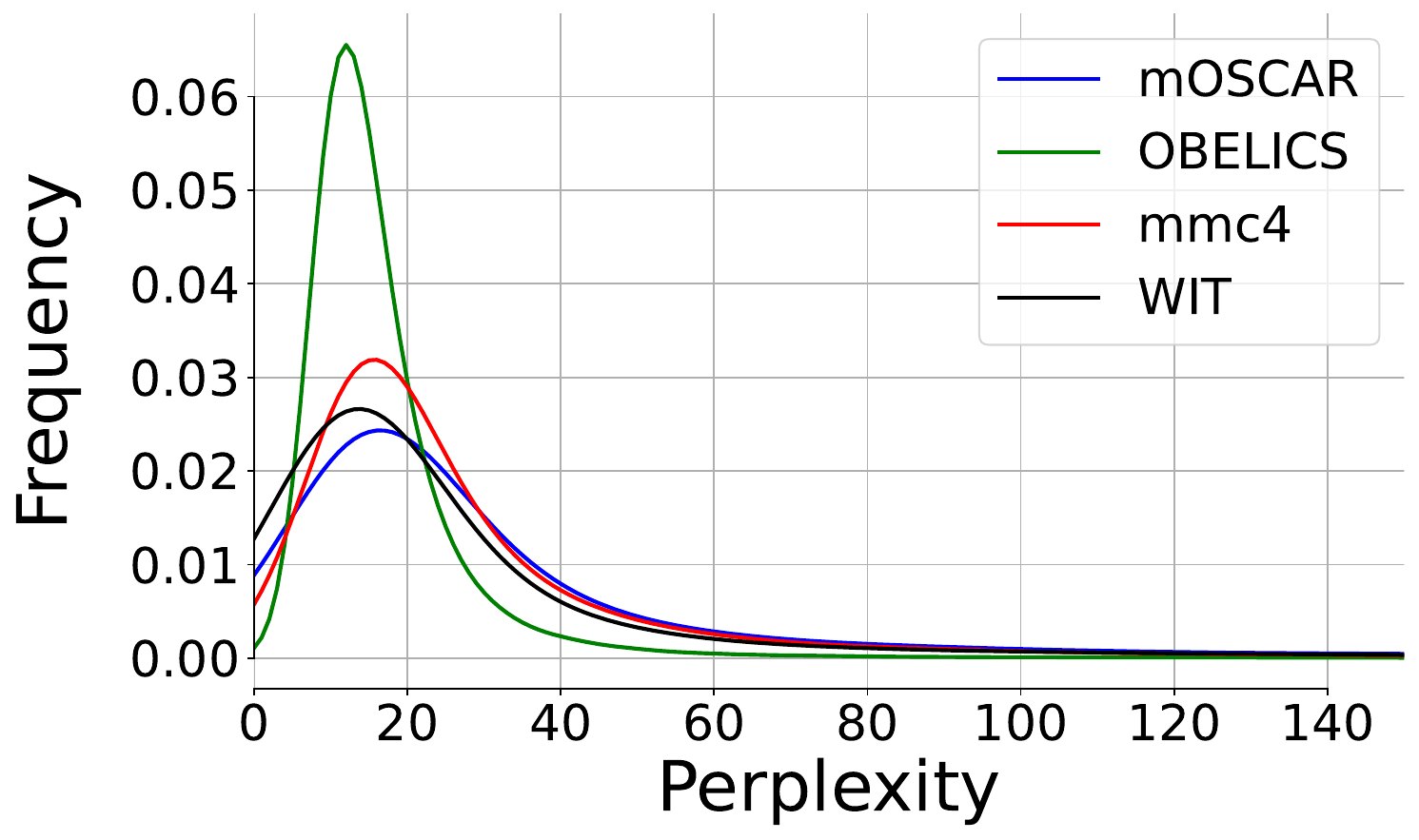}
  \caption{Perplexity of 100K random documents from different datasets.}
  \label{fig:ppl-quality}
\end{figure}

\begin{table}[!h]
\centering\small
\begin{tabular}{lrr}
\toprule
     &  Vendi score & Dist. $n$-gram ratio  \\ 
\midrule
mOSCAR & \underline{69.05} ($\pm$ 0.14) & 0.472 ($\pm$ 0.002) \\
mmc4 & 67.93 ($\pm$ 0.12) & \underline{0.494} ($\pm$ 0.002) \\
OBELICS & 58.49 ($\pm$ 0.09) & 0.488 ($\pm$ 0.001) \\
WIT & \textbf{73.30} ($\pm$ 0.09) &  \textbf{0.530} ($\pm$ 0.001) \\
\bottomrule
\end{tabular} 
\caption{Average text diversity scores ($\pm$ standard error) of text documents.}
\label{tab:datasets-text-diversity}
\end{table}

For content diversity, we randomly sample 30M documents for mOSCAR, mmc4 and OBELICS and 3M documents for WIT and represent the documents by their SimCSE embedding. We compute the Vendi Score with cosine similarity on a randomly sampled subset of 65,536 documents. Table~\ref{tab:datasets-text-diversity} shows that mOSCAR English content is more diverse than mmc4 and OBELICS but less diverse than WIT.
For lexical diversity, we randomly sample 3M documents for mOSCAR, mmc4, OBELICS and WIT and compute the distinct $n$-gram ratio on a subset of 8,192 documents for $n$ from 1 to 4. Table~\ref{tab:datasets-text-diversity} shows that mOSCAR is slightly less lexically diverse than OBELICS and mmc4, while WIT is by far the most diverse.

\paragraph{Quality}
To evaluate document quality, we focus on English documents and compute their perplexity using Gemma-2B \citep{team2024gemma}. Figure~\ref{fig:ppl-quality} shows the kernel density estimation of the distribution of the perplexity of 100K randomly sampled documents from different datasets: mOSCAR is comparable to mmc4 and WIT, while OBELICS appears to be the of the highest quality. mOSCAR is therefore comparable to other interleaved image-text dataset in terms of quality and diversity of its English subset. It is however more diverse than English-only datasets by its multilingual construction and more than 10 times larger than existing multilingual interleaved image-text datasets such as WIT.


\subsubsection{Image diversity}

\paragraph{Comparison with other datasets}


\begin{table}[!h]
\centering\small
\begin{tabular}{ccc}
\toprule
   mOSCAR    &   LAION-400M & WIT \\ 
\midrule
\underline{55.74} ($\pm$ 0.16)  &  \textbf{67.59} ($\pm$ 0.16) & 36.14 ($\pm$ 0.08)    \\ \bottomrule
\end{tabular}
\caption{Average Vendi score ($\pm$ standard error) of images sampled from different datasets.}
\label{tab:datasets-img-diversity}
\end{table}

We compute the Vendi Score on random samples of images for different datasets, comparing the images from English mOSCAR documents with those from Conceptual Captions \citep{changpinyo2021conceptual}, LAION-400M \citep{schuhmann2021laion} and WIT \citep{srinivasan2021wit}. We represent each image by its SigLIP\footnote{We use \texttt{siglip-base-patch16-224}.} \citep{siglip} embedding 
and compute the Vendi score on batches of size 65,536 and a total of 1M images for each dataset. In Table~\ref{tab:datasets-img-diversity}, we notice that the set of images in mOSCAR documents are more diverse than images from WIT documents but less diverse than LAION-400M.


\begin{table}[!h]
\centering\small
\begin{tabular}{cc}
\toprule
   English    &   All  \\ 
\midrule
52.36 ($\pm$ 0.18)  &  \textbf{54.78} ($\pm$ 2.29)     \\
\bottomrule
\end{tabular}
\caption{Average Vendi score ($\pm$ standard error) of images sampled from
mOSCAR (English vs. any language).}
\label{tab:multilingual-img-diversity}
\end{table}

\paragraph{Multilingual diversity}

We also compare the diversity of images from English documents and of images sampled from documents of any language (English included). We use multilingual SigLIP \citep{chen2023pali} trained on WebLI \citep{chen2022pali} to compute image embeddings used to get the Vendi score. We again use a batch of size 65,536 and a total of 3M images, and we do not sample multiple images from a same document. For the multilingual setting, we randomly sample 50 languages and an equal number of images for each language to build the batch. As we did not do any image deduplication across languages, we could expect to have less diversity in the multilingual setting. 
However, Table~\ref{tab:multilingual-img-diversity} shows that the set of images is on average more diverse when sampled from all documents than from English-only documents. This means that the distribution of images is not exactly the same across languages, potentially due to cultural differences.




\begin{table*}[!th]
    \centering\small
    \resizebox{.9\textwidth}{!}{
    \begin{tabular}{lccccccccc} \toprule
       &   & \multirow{2}{*}{xFlickR\&CO} & \multirow{2}{*}{XM3600} & \multirow{2}{*}{xGQA} & \multirow{2}{*}{MaXM} & \multirow{2}{*}{MaRVL} & \multirow{2}{*}{XVNLI} & \multirow{2}{*}{Multi30K} & \multirow{2}{*}{CoMMuTE} \\
       & \#shots &  & & & & & & & \\
        \midrule
        
     \multirow{3}{*}{Multi. OF}   & 0 & 16.91 & 7.45 & 26.95 & 22.33 & 49.56 & 33.88 & 22.91 & 63.34 \\
        & 4 & 34.80 & 22.18 & 32.33 & 26.33 & 49.64 & 34.07 & 23.27 & 63.22 \\
       \textit{mOSCAR + cap.} & 8 & 36.90 & 23.48 & 34.24 & 27.08 & \textbf{51.48} & \textbf{36.60} & 23.59 & \textbf{63.54} \\
        & 16 & \textbf{39.46} & \textbf{23.67} & \textbf{35.23} & \textbf{27.47} & 49.84 & 34.85 & \textbf{23.85} & 62.78 \\ \midrule
        
    \multirow{3}{*}{Multi. OF}   & 0 & 9.57 & 4.21 & 8.62 & 4.01 & \textbf{49.88} & 33.76 & \hphantom{0}0.00 & 61.36 \\
        & 4 & 13.20 & 9.26 & 13.45 & 4.15 & 49.54 & 32.04 & \hphantom{0}0.00 & 61.13 \\
     \textit{cap. only}   & 8 & 18.00 & 10.35 & 12.82 & 4.88 & 49.65 & 33.71 & \hphantom{0}0.01 & 60.90 \\
        & 16 & 19.87 & 12.07 & 13.37 & 4.89 & 49.79 & 32.70 & \hphantom{0}0.74 & 60.25 \\ \bottomrule
        \end{tabular}
        }
    \caption{Results averaged over all languages. Multi. OF refers to multilingual Open Flamingo, \textit{mOSCAR + cap.} refers to the model trained on text-image pairs and mOSCAR while \textit{cap. only} refers to the model trained only on text-image pairs. \textbf{Bold} is best result.}
    \label{tab:results-tab}
\end{table*}

\begin{table*}[!th]
    \centering\small
    \resizebox{.9\textwidth}{!}{
    \begin{tabular}{lccccccccc} \toprule
       &   & \multirow{2}{*}{xFlickR\&CO} & \multirow{2}{*}{XM3600} & \multirow{2}{*}{xGQA} & \multirow{2}{*}{MaXM} & \multirow{2}{*}{MaRVL} & \multirow{2}{*}{XVNLI} & \multirow{2}{*}{Multi30K} & \multirow{2}{*}{CoMMuTE} \\
       & \#shots &  & & & & & & & \\
        \midrule
        
     \multirow{3}{*}{Multi. OF (35M)}   & 0 & 19.07 & 8.73 & 25.08 & 19.64 & \textbf{49.77} & 33.01 & 22.70 & \textbf{63.75} \\
        & 4 & 34.32 & 20.59 & 31.90 & 23.90 & 49.67 & 36.07 & 22.79 & 63.65 \\
       \textit{mOSCAR + cap.} & 8 & 36.77 & 22.15 & 33.9 & 24.41 & 49.72 & \textbf{37.16} & 23.21 & 63.00 \\
        & 16 & \textbf{37.63} & \textbf{22.24} & \textbf{35.71} & \textbf{25.38} &  49.73 & 35.36 & 23.48 & 62.77 \\ \midrule
        
    \multirow{3}{*}{Multi. OF (35M)}   & 0 & 9.39 & 4.67 & 19.81 & 14.63 & 49.71 & 32.78 & \textbf{26.99} & 56.75 \\
        & 4 & 7.68 & 2.99 &  25.68 & 16.12 & 49.72 & 33.51 & \textbf{26.99} & 53.27 \\
     \textit{WIT + cap.}   & 8 & 8.91 & 3.63 & 27.06 & 16.81 & 49.74 & 32.77 & \textbf{26.99} & 55.33 \\
        & 16 & 9.74	& 4.14 & 28.14 & 16.34 & 49.74 & 33.63 & \textbf{26.99} & 54.04 \\ \bottomrule
        \end{tabular}
        }
    \caption{Results averaged over all languages and comparison between a model trained on WIT and a checkpoint of multilingual Open Flamingo trained on 35M mOSCAR documents (full model was trained on 50M mOSCAR documents). Both models were trained on 35M documents from their respective training datasets and 70M text-image pairs for fair comparison. Multi. OF (35M) refers to multilingual Open Flamingo trained on 35M documents. \textbf{Bold} is best result.}
    \label{tab:results-tab-wit}
\end{table*}

\section{Training a multilingual multimodal language model}

We train a multilingual Flamingo-like model on mOSCAR that we call multilingual Open Flamingo. As adding captioning data to training data has been shown to improve zero-shot performance \citep{mckinzie2024mm1}, we additionally train on LAION-400M, which we re-captioned using BLIP \citep{li2022blip}, filtered with CLIP score \citep{radford2021learning} and translated using distilled NLLB-600M \citep{visheratin2023nllb} following the proportion of languages found in mOSCAR. We use Gemma-2B \citep{team2024gemma} as the underlying language model and we train the model on 50M mOSCAR documents and 100M randomly sampled image-text pairs. We also train a model on 300M image-text pairs, a model trained on 35M WIT \citep{srinivasan2021wit} documents and 70M text-image pairs and a model trained on 50M mOSCAR documents from the English subset and 100M English image-text pairs as comparison baselines. We additionally compare with OpenFlamingo-3B-MPT \citep{awadalla2023openflamingo} as the \textit{translate-test} baseline. The full list of languages for training and the implementation details can be found in Appendix~\ref{sec:impl-details}.


\begin{table*}[!h]
    \centering
    \resizebox{.8\textwidth}{!}{\begin{tabular}{lccccccc} \toprule
    & \#shots   & xFlickR\&CO & XM3600 & xGQA & MaXM & XVNLI \\ \midrule
 \multirow{3}{*}{Multilingual OF} & 0 & 29.64 & 42.57 & 34.24 & 36.58 & 34.62 \\ 
 & 4 & 51.47 & 77.98 & 37.91 & 38.13 & 33.59 \\
 \textit{mOSCAR + cap.} & 8 & 56.75 & 77.64 & 39.44 & \textbf{38.52} & \textbf{38.75} \\
 & 16 & \textbf{59.89} & 78.18 & 40.09 & 35.80 & 36.60 \\ \midrule

 \multirow{3}{*}{English OF} & 0 & 32.70 & 43.75 & 34.71 & 36.19 & 35.82 \\ 
 & 4 & 51.39 & 75.33 & 37.48 & 37.96 & 34.88 \\
\textit{English mOSCAR + English cap.} & 8 & 51.44 & 77.73 & 39.64 & 38.35 & 36.86 \\
 & 16 & 59.24 & \textbf{78.38} & \textbf{40.36} & 37.35 & 37.11 \\ \bottomrule 
    \end{tabular}}
    \caption{Results on the English subsets of the test sets and comparison between multilingual Open Flamingo and an Open Flamingo trained on the English subset of mOSCAR and English text-image pairs (English OF). Both models were trained on 50M documents from their respective training datasets and 100M text-image pairs for fair comparison. \textbf{Bold} is best result.}
    \label{tab:english-vs-multilingual}
\end{table*}

\subsection{Evaluation setup} \label{sec:eval}

We evaluate the models using a broad set of image-text multilingual tasks and benchmarks. We use the IGLUE benchmark \citep{bugliarello2022iglue} composed of XVNLI, MaRVL \citep{liu-etal-2021-visually} to test reasoning, xGQA \citep{pfeiffer-etal-2022-xgqa} to test visual question answering capabilities and xFlickr\&CO \citep{young-etal-2014-image, karpathy2015deep, yoshikawa-etal-2017-stair} for captioning. We also include Crossmodal-3600 (XM3600) \citep{thapliyal-etal-2022-crossmodal} and MaXM \citep{changpinyo2022maxm} as they cover a broader range of languages. To test to what extent models trained on mOSCAR can perform zero-shot multimodal machine translation (MMT), we also test on Multi30K \citep{W16-3210, elliott-EtAl:2017:WMT, barrault2018findings} and CoMMuTE \citep{futeral-etal-2023-tackling}. For captioning we compute the CideR \citep{vedantam2015cider} score and we tokenize references and model outputs with the Stanford Core NLP tokenizer for English and Stanza \citep{qi2020stanza} tokenizers for other languages. To evaluate Multi30k, we compute BLEU \citep{papineni-etal-2002-bleu} score from Sacrebleu \citep{post-2018-call} with \textit{13a} tokenization and default parameters. We use accuracy for CoMMuTE. More details can be found in Appendix~\ref{sec:eval-details}.

\begin{table}[!h]
    \centering
    \resizebox{.48\textwidth}{!}{
    \begin{tabular}{lccccc} \toprule
       &  & \multirow{2}{*}{xGQA} & \multirow{2}{*}{MaXM} & \multirow{2}{*}{MaRVL} & \multirow{2}{*}{XVNLI} \\
       & \#shots & & & & \\
        \midrule
    \multirow{4}{*}{OF-3B MPT}   & 0  & 18.34 & 7.68 & 49.75 & 32.73 \\
        & 4  & 22.97 & 7.82 & 49.70 & 35.82 \\
      & 8 & 28.57 & 8.32 & 49.71 & 31.29 \\
        & 16  & 31.82 & 9.04 & 49.72 & 33.29 \\ \midrule
        
    \multirow{3}{*}{Multi. OF}   & 0 & 30.16 & 10.06 & 49.93 & 34.66 \\
        & 4 & 35.55 & 9.89 & 48.99 & 36.10 \\
     \textit{mOSCAR + cap.}   & 8 & 36.78 & 10.12  & \textbf{50.54} & \textbf{39.69} \\
        & 16 & \textbf{37.75} & \textbf{11.49} & 49.57 & 37.97 \\ \bottomrule
    \end{tabular}}
    \caption{\textit{Translate-test} results averaged over languages where all benchmarks were translated from local languages into English using Google Translate API. Multi. OF \textit{mOSCAR + cap.} refers to Multilingual Open Flamingo trained on mOSCAR and text-image pairs while OF-3B MPT refers to Open Flamingo \citep{awadalla2023openflamingo} based on MPT \citep{MosaicML2023Introducing} and trained on mmc4 \citep{mmc4} and English text-image pairs.}
    \label{tab:results-tab-translate-test}
    \end{table}

\begin{figure}[!th]
    \centering
    \includegraphics[width=0.8\linewidth]{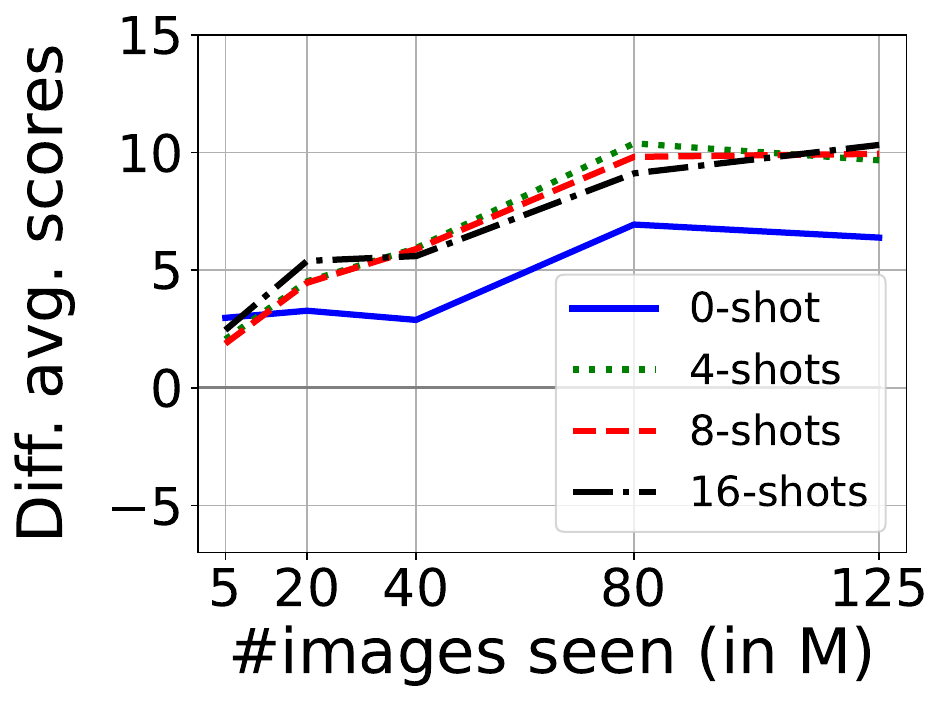}
    \caption{Score differences averaged over benchmarks and languages between the model trained on mOSCAR + text-image pairs and the model trained only on text-image pairs. \textbf{Bold} is best result.}
    \label{fig:scores-diff}
\end{figure}

\subsection{Results} 

\Cref{tab:results-tab,tab:results-tab-translate-test} show the average results across all languages. Full results are available in Appendix~\ref{sec:detailed-res}. We notice that the multilingual OpenFlamingo trained additionally on mOSCAR gets better results than the model trained on captioning data only while having seen fewer image-text pairs during training. More importantly, when increasing the number of few-shot examples from 0 to 16, it sees gains of on average +6.71 points on VQA benchmarks  and +19.39 CideR points on captioning benchmarks. In contrast, the model trained on text-image pairs only sees gains of +2.82 and +9.08 points respectively. In cross-modal machine translation, the model additionally trained on interleaved data is again much better than the one trained on just captioning data, which is not able to translate the Multi30k benchmark at all.\footnote{Most of the time, the model is not able to follow the prompt and only outputs the end of sequence token.} Moreover, mOSCAR helps the model to learn to zero-shot disambiguate translations as shown by the improved average score on CoMMuTE (63.54) compared to the model trained on captions only (61.36).

Multilingual Open Flamingo trained on mOSCAR \& text-image pairs is also better than OpenFlamingo 3B MPT evaluated on translate test benchmarks\footnote{This means benchmarks were translated from local languages to English, using Google Translate API}. However, we obtain the best results (except for MaXM) by evaluating our multilingual Open Flamingo on the translate-test benchmarks since the underlying language model (Gemma-2B) is far better in English than other languages. We also notice that all models struggle with reasoning classification tasks (MaRVL, XVNLI) where they obtain scores close to random guessing.

Table~\ref{tab:results-tab-wit} additionally shows that Multilingual Open Flamingo trained on mOSCAR obtains much better results than the same model trained on WIT for equivalent training data seen during training\footnote{We select the checkpoint of multilingual Open Flamingo trained on 35M documents and 70M captions to have fair comparison.} (except for Multi30K benchmark) which means mOSCAR is better suited than WIT for training multilingual mLLMs. Eventually, Table~\ref{tab:english-vs-multilingual} shows that we don't face a drop in performances in English performances when training the model on 43 languages (multilingual Open Flamingo) in comparison to training it on the English subset of mOSCAR and English text-image pairs.

Additional comparison results with InternVL2 \citep{chen2024far}, Llava-NeXT \citep{li2024llavanext-strong}, PaliGemma \citep{beyer2024paligemma} and Idefics2 \citep{laurenccon2024matters} can be found in Appendix~\ref{sec:res-sota}.

We additionally compare results at different training steps, defined by the number of images seen during training. Figure~\ref{fig:scores-diff} shows the difference of averaged scores between the model trained on all data and the model trained only on text-images pairs. We notice that the gap first decreases until 20M images seen and keep increasing over training at all training steps after that. Particularly, the gap is wider for few-shot learning.

\section{Conclusion}\label{sec:conclusion}

We introduced mOSCAR, a large-scale multilingual and multimodal dataset covering 163 languages and composed of 303M documents, 200B tokens and 1.15B images. We showed that mOSCAR is of good quality, diverse and could be used to train a multilingual and multimodal LLM. We also proved that training on mOSCAR led to the emergence of in-context learning capabilities in several languages. We eventually ensured that mOSCAR is as safe as possible by applying a series of filtering steps to remove NSFW and toxic content.

\section*{Limitations}
We did not conduct any analysis of the biases of mOSCAR as this is challenging in a multilingual setting. As it is crawled from the Internet, it is indeed possible that mOSCAR reflects biases widespread on it. Training a model on mOSCAR must therefore be combined with additional alignment training steps to mitigate potential biases towards groups of people. Nevertheless, by its multilingual nature, mOSCAR is a step towards the inclusion of more languages, cultures, and people in accessing mLLMs.

\section*{Acknowledgements}
This work was granted access to the HPC resources
of IDRIS under the allocation 2024-AD011014232R1, 2023-AD011014232 and 2023-AD011012254 made by GENCI. It was
also partly funded by the last three authors’ chairs
in the PRAIRIE institute funded by the French national agency ANR as part of the “Investissements
d’avenir” programme under the reference ANR-19-
P3IA-0001. We deeply thanks the Jean-Zay support team. We also thank Filip Šedivý for insightful discussions regarding the removal of CSAM, Thorn for having provided access to their CSAM detector, Zeeshan Khan for discussions regarding the training of the models and Victoria Le Fourner for having manually checked subsamples of NSFW images.
\bibliography{custom}
\bibliographystyle{acl_natbib}


\clearpage

\appendix

\section{Appendix}

\subsection{mOSCAR languages \& statistics} \label{sec:langs-stats}
Table~\ref{langs_stats} shows the number of documents, the number of images, and the number of NLLB tokens for each language of mOSCAR.

\subsection{Topic modeling} \label{sec:topic-modeling}
We analyzed the topic found in different language subsets of mOSCAR with BERTopic \citep{grootendorst2022bertopic}. We first preprocessed $\sim$30K randomly sampled documents in tokens using stanza tokenizers \citep{qi2020stanza}, we then represent each document with their document embedding obtained with a multilingual sentence-bert\footnote{\texttt{sentence-transformers/paraphrase-multilingual-MiniLM-L12-v2}} \citep{reimers-2019-sentence-bert} before running BERTopic to cluster the documents. \Cref{topic-mode-en,topic-mode-als,topic-mode-ar,topic-mode-el,topic-mode-hi} show some of the most salient English-translated strings representing each cluster and their weight (in percentage) in the subsets of mOSCAR. While many clusters are shared between subsets, we can observe that some of them are culturally-specific (i.e. they are related to the culture of the language subset and only appears on this subset). For instance, Table~\ref{topic-mode-hi} shows topics of the Hindi subset and we observe a cluster related to Hindu gods described by the words `ganesha', `shiva' or `shani dev'. We also observe a cluster related to festivals hosted in India, it is described by the words `festival', `dussehra' or `janmashtami'. We can also observe that among clusters shared between several language subsets, some contain knowledge culturally-specific as exemplified by the cluster related to jewelry in the Hindi subset which contain the word `mangalsutra' which is a jewel worn by men during weddings in India. Eventually, we can observe that the English subset contains a strong cluster related to the Christian religion. While a similar cluster appears in the Tosk Albanian subset, it is related to Islam. This shows the importance of a multilingual corpora to have data representing more cultures around the world.

\subsection{Examples of documents} \label{sec:ex-docs}
\begin{figure*}[!htbp]
    \centering
\includegraphics[width=.9\textwidth]{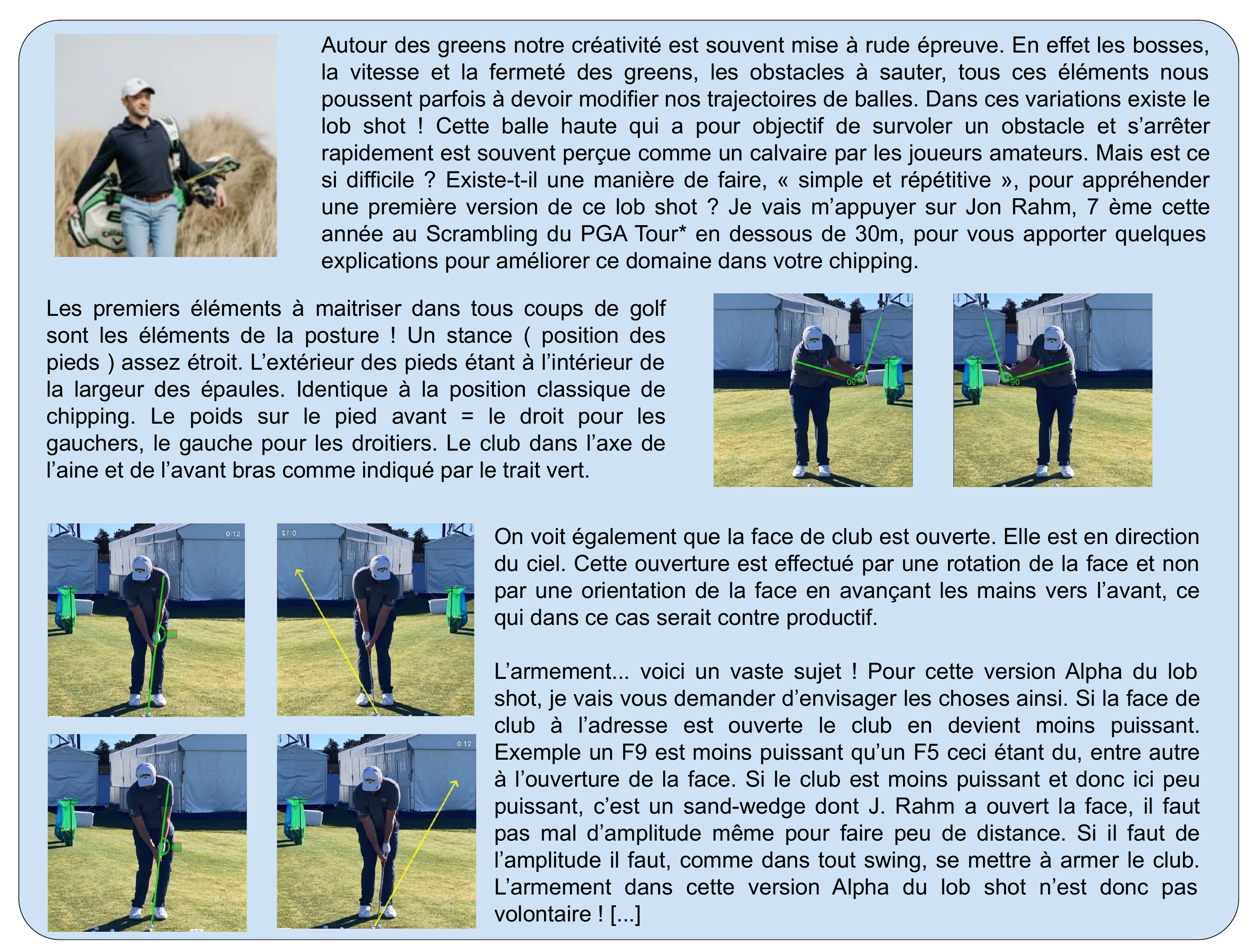}
    \caption{Example of a French document.}
    \label{fig:eg-golf-french}
\end{figure*}

\begin{figure*}[!htbp]
    \centering
\includegraphics[width=\textwidth]{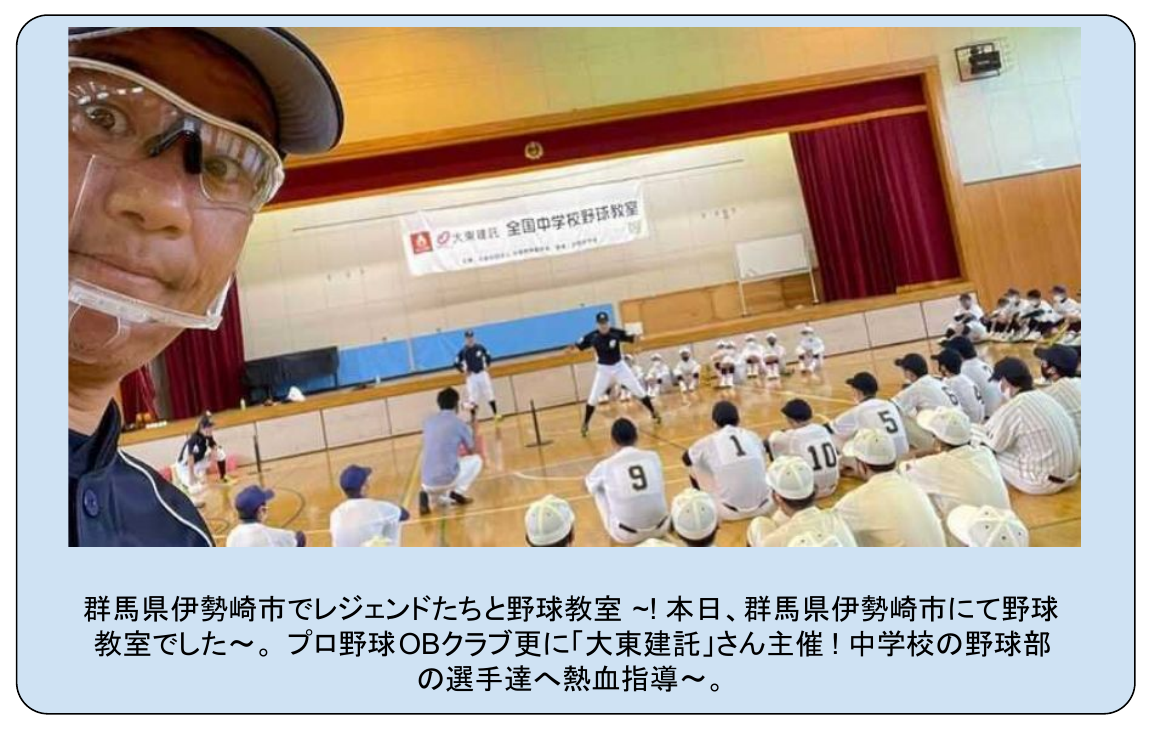}
    \caption{Example of a Japanese document.}
    \label{fig:eg-baseball-jpn}

\vspace{1cm}
    \centering
\includegraphics[width=\textwidth]{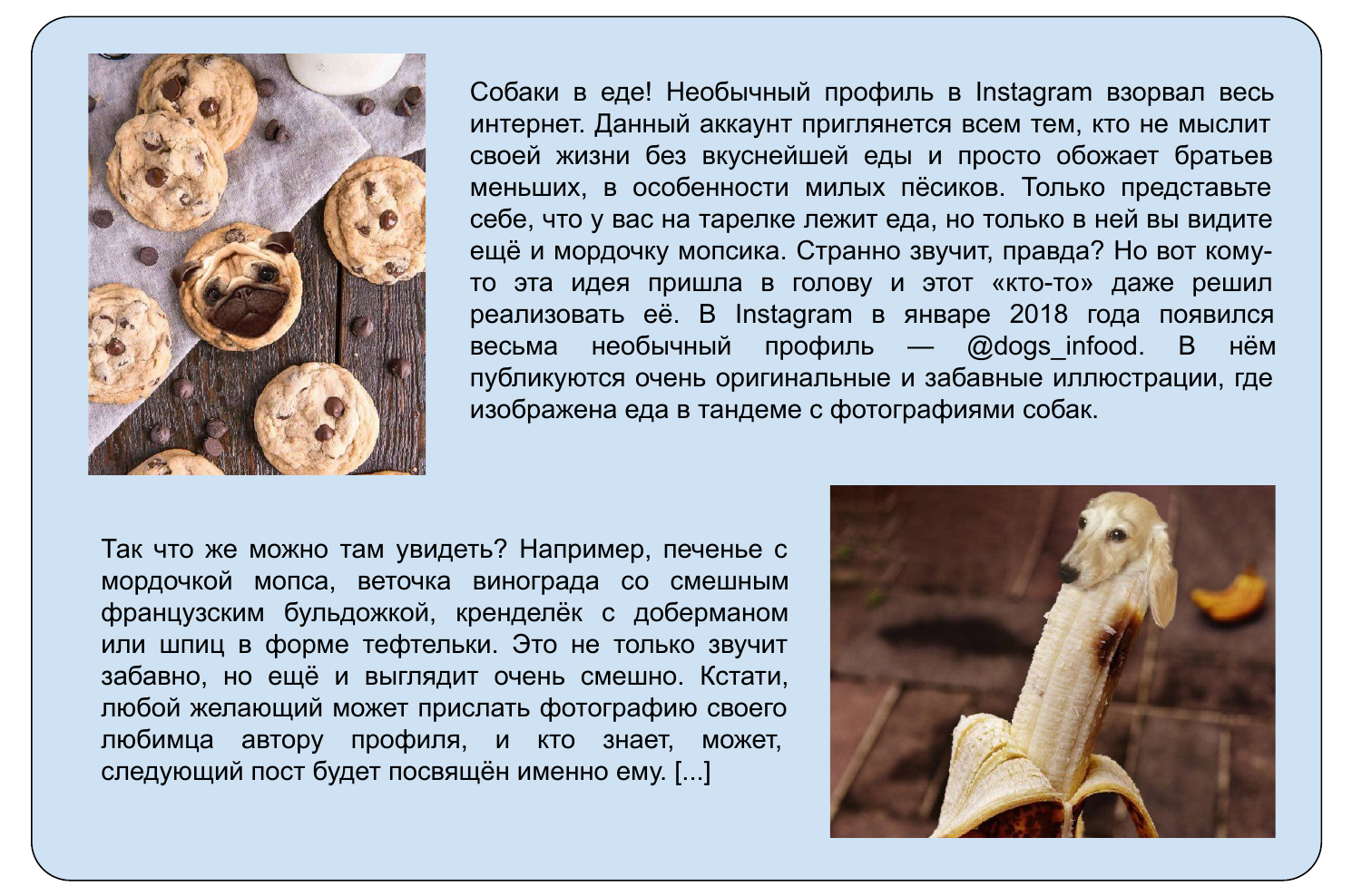}
    \caption{Example of a Russian document.}
    \label{fig:eg-doc-rus}
\end{figure*}

\begin{figure*}[!htbp]
    \centering
\includegraphics[width=\textwidth]{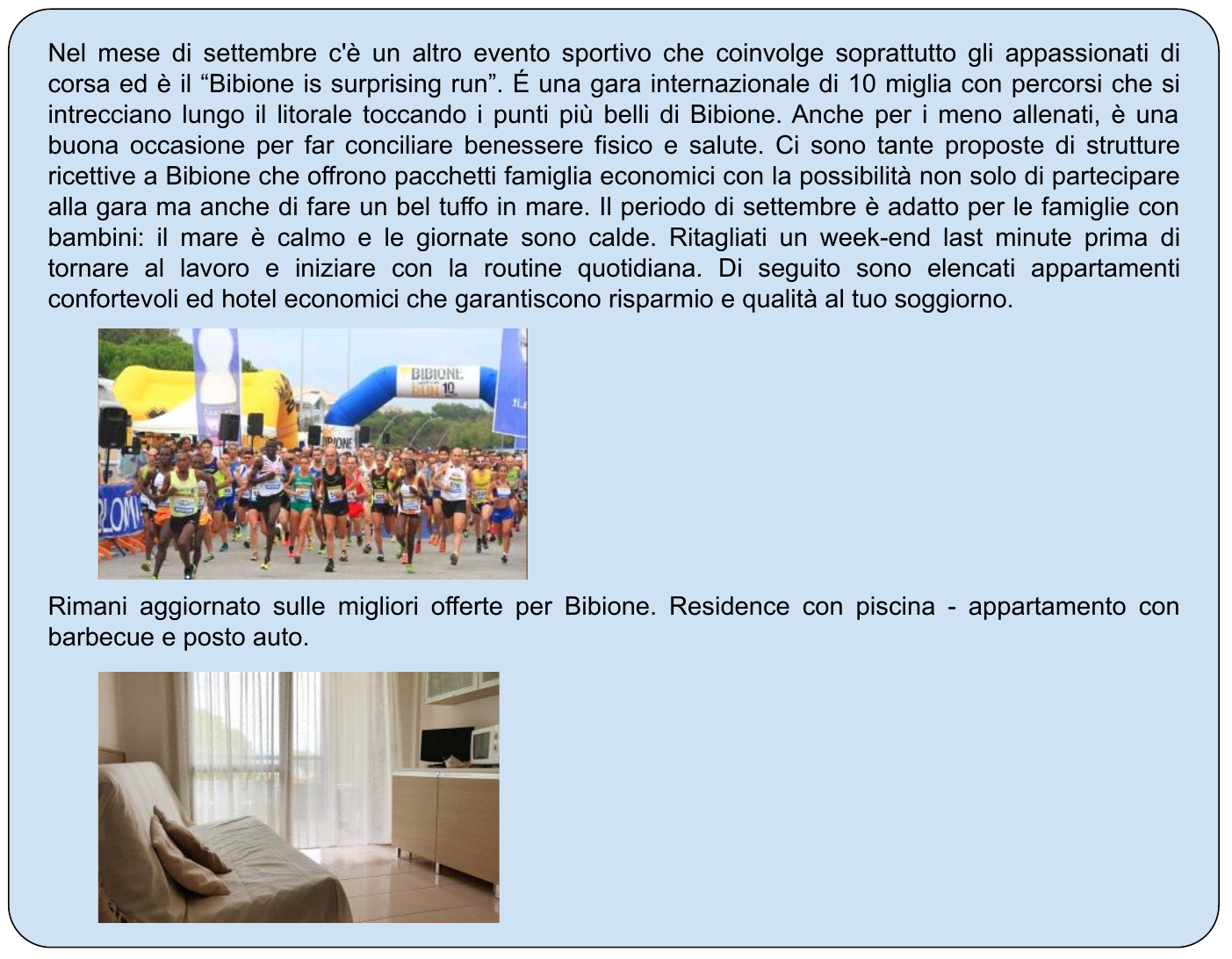}
    \caption{Example of an Italian document.}
    \label{fig:eg-doc-ita}
\end{figure*}

\begin{figure*}[!htbp]
    \centering
\includegraphics[width=\textwidth]{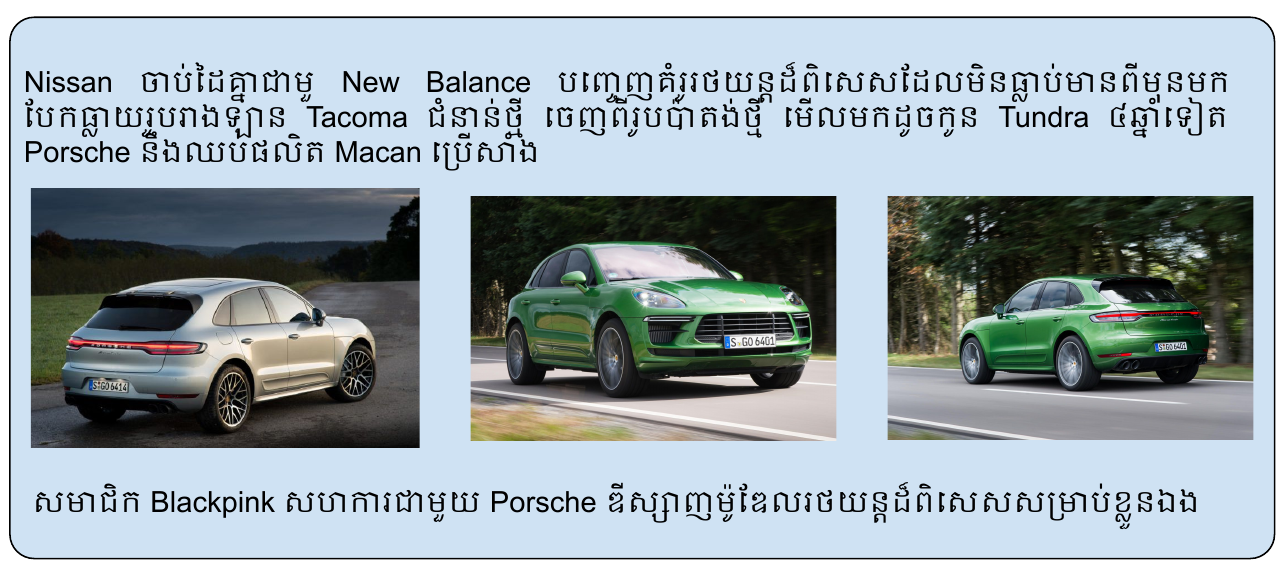}
    \caption{Example of a Khmer document.}
    \label{fig:eg-doc-khm}
\end{figure*}

\begin{figure*}[!htbp]
    \centering
\includegraphics[width=\textwidth]{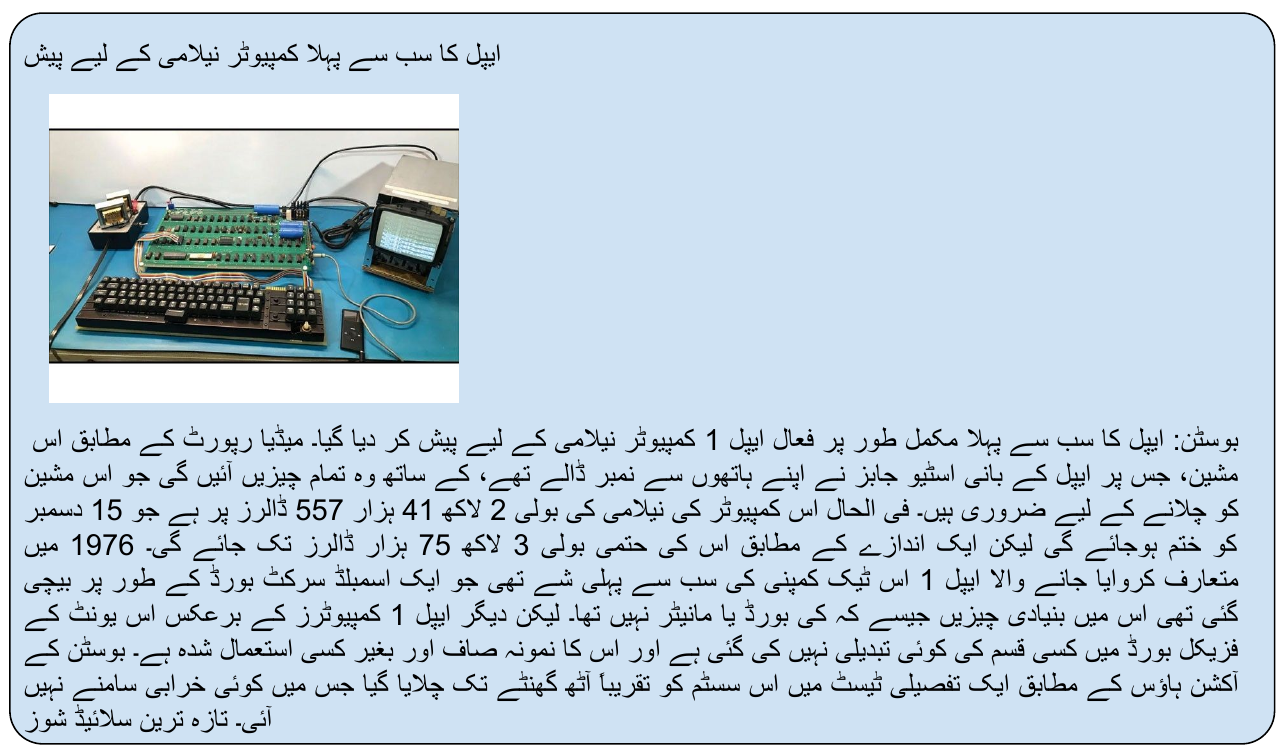}
    \caption{Example of an Urdu document.}
    \label{fig:eg-doc-urd}
\end{figure*}

\Cref{fig:eg-golf-french,fig:eg-baseball-jpn,fig:eg-doc-rus,fig:eg-doc-ita,fig:eg-doc-khm,fig:eg-doc-urd} provide examples of documents found in mOSCAR for different languages. We can observe the diversity in terms of content of the documents. Figure~\ref{fig:eg-golf-french} is a French document describing a specific shot of golf illustrated by images which provide very detailed description of images. Figure~\ref{fig:eg-doc-rus} shows a Russian document describing an Instagram account with pictures of food with animal heads which is reminiscent of a famous problem of computer vision consisting in classifying images between chihuahuas and muffins.

\onecolumn
{\small\tabcolsep=3pt  
\begin{longtable}[c]{llllcrrr} 
     \toprule
    \multicolumn{4}{c}{Languages} && \multicolumn{3}{c}{Statistics} \\ \cmidrule{1-4} \cmidrule{6-8} 
       Lang. name  & Code & Family & Script && \#documents & \#images & \#tokens  \\ \midrule
       \endfirsthead

    \toprule
    \multicolumn{4}{c}{Languages} && \multicolumn{3}{c}{Statistics} \\ \cmidrule{1-4} \cmidrule{6-8} 
       Lang. name  & Code & Family & Script && \#documents & \#images & \#tokens  \\ \midrule
       \endhead

       \bottomrule
        \endfoot
        \bottomrule \\
        \caption{Languages \& Statistics}
        \label{langs_stats}
        \endlastfoot
       
Acehnese & \texttt{ace\_Latn} & Austronesian & Latin && 2,159 & 9,026 & 1,395,381 \\
 Mesopotamian Arabic & \texttt{acm\_Arab} & Afro-Asiatic & Arabic && 1,282 & 5,621 & 704,549 \\
 Tunisian Arabic & \texttt{aeb\_Arab} & Afro-Asiatic & Arabic && 5,933 & 34,270 & 2,308,455 \\
 Afrikaans & \texttt{afr\_Latn} & Indo-European & Latin && 50,061 & 211,876 & 38,761,504 \\
 South Levantine Arabic & \texttt{ajp\_Arab} & Afro-Asiatic & Arabic && 8,603 & 69,051 & 3,869,688 \\
 Tosk Albanian & \texttt{als\_Latn} & Indo-European & Latin && 856,144 & 2,543,758 & 441,244,377 \\
 Amharic & \texttt{amh\_Ethi} & Afro-Asiatic & Ge‘ez && 39,031 & 149,739 & 33,768,732 \\
 North Levantine Arabic & \texttt{apc\_Arab} & Afro-Asiatic & Arabic && 16,198 & 110,792 & 8,268,237 \\
 Modern Standard Arabic & \texttt{arb\_Arab} & Afro-Asiatic & Arabic && 3,794,792 & 14,757,353 & 3,346,786,610 \\
 Najdi Arabic & \texttt{ars\_Arab} & Afro-Asiatic & Arabic && 52,102 & 261,275 & 39,066,487 \\
 Moroccan Arabic & \texttt{ary\_Arab} & Afro-Asiatic & Arabic && 117,957 & 584,301 & 188,462,338 \\
 Egyptian Arabic & \texttt{arz\_Arab} & Afro-Asiatic & Arabic && 761,113 & 3,785,164 & 635,018,784 \\
 Assamese & \texttt{asm\_Beng} & Indo-European & Bengali && 2,947 & 7,228 & 543,676 \\
 Asturian & \texttt{ast\_Latn} & Indo-European & Latin && 87,649 & 533,723 & 25,499,269 \\
 Awadhi & \texttt{awa\_Deva} & Indo-European & Devanagari && 8,179 & 29,142 & 2,293,620 \\
 Central Aymara & \texttt{ayr\_Latn} & Aymaran & Latin && 10,112 & 57,294 & 2,343,403 \\
 South Azerbaijani & \texttt{azb\_Arab} & Turkic & Arabic && 3,411 & 14,825 & 3,143,946 \\
 North Azerbaijani & \texttt{azj\_Latn} & Turkic & Latin && 511,832 & 1,796,046 & 256,160,442 \\
 Bashkir & \texttt{bak\_Cyrl} & Turkic & Cyrillic && 3,287 & 12,031 & 2,600,135 \\
 Bambara & \texttt{bam\_Latn} & Manding & Latin && 3,011 & 17,666 & 446,961 \\
 Balinese & \texttt{ban\_Latn} & Austronesian & Latin && 787 & 4,894 & 392,978 \\
 Belarusian & \texttt{bel\_Cyrl} & Indo-European & Cyrillic && 60,443 & 276,672 & 71,854,171 \\
 Bemba & \texttt{bem\_Latn} & Atlantic–Congo & Latin && 582 & 3,018 & 1,021,026 \\
 Bengali & \texttt{ben\_Beng} & Indo-European & Bengali && 204,475 & 758,222 & 30,400,395 \\
 Bhojpuri & \texttt{bho\_Deva} & Indo-European & Devanagari && 4,190 & 18,339 & 715,786 \\
 Banjar & \texttt{bjn\_Latn} & Austronesian & Latin && 1,764 & 9,017 & 1,093,443 \\
 Bosnian & \texttt{bos\_Latn} & Indo-European & Latin && 635,750 & 2,642,491 & 423,073,661 \\
 Buginese & \texttt{bug\_Latn} & Austronesian & Latin && 584 & 2,379 & 167,459 \\
 Bulgarian & \texttt{bul\_Cyrl} & Indo-European & Cyrillic && 2,578,191 & 11,601,214 & 1,736,106,287 \\
 Catalan & \texttt{cat\_Latn} & Indo-European & Latin && 1,132,056 & 4,638,966 & 598,942,711 \\
 Cebuano & \texttt{ceb\_Latn} & Austronesian & Latin && 14,924 & 75,258 & 10,221,371 \\
 Czech & \texttt{ces\_Latn} & Indo-European & Latin && 3,736,126 & 12,683,461 & 2,767,295,966 \\
 Central Kurdish & \texttt{ckb\_Arab} & Indo-European & Arabic && 36,413 & 135,461 & 21,622,335 \\
 Crimean Tatar & \texttt{crh\_Latn} & Turkic & Latin && 2,744 & 10,079 & 1,173,321 \\
 Welsh & \texttt{cym\_Latn} & Indo-European & Latin && 38,616 & 155,591 & 27,237,252 \\
 Danish & \texttt{dan\_Latn} & Indo-European & Latin && 2,020,516 & 9,214,031 & 1,207,829,704 \\
 German & \texttt{deu\_Latn} & Indo-European & Latin && 20,265,504 & 86,393,702 & 8,315,212,019 \\
 Southwestern Dinka & \texttt{dik\_Latn} & Nilo-Saharan & Latin && 1,233 & 4,766 & 1,098,795 \\
 Greek & \texttt{ell\_Grek} & Indo-European & Greek && 4,895,433 & 15,147,284 & 2,909,427,055 \\
 English & \texttt{eng\_Latn} & Indo-European & Latin && 51,658,029 & 205,363,181 & 32,599,001,993 \\
 Esperanto & \texttt{epo\_Latn} & Artificial & Latin && 23,619 & 112,577 & 26,976,847 \\
 Estonian & \texttt{est\_Latn} & Uralic & Latin && 1,022,368 & 5,108,102 & 589,045,973 \\
 Basque & \texttt{eus\_Latn} & Isolate & Latin && 682,599 & 2,914,120 & 259,930,954 \\
 Faroese & \texttt{fao\_Latn} & Indo-European & Latin && 14,921 & 56,934 & 6,579,921 \\
 Fijian & \texttt{fij\_Latn} & Austronesian & Latin && 1,039 & 4,039 & 416,670 \\
 Finnish & \texttt{fin\_Latn} & Uralic & Latin && 2,377,155 & 10,263,171 & 1,749,904,041 \\
 French & \texttt{fra\_Latn} & Indo-European & Latin && 19,963,542 & 76,851,982 & 13,818,099,493 \\
 Friulian & \texttt{fur\_Latn} & Indo-European & Latin && 15,823 & 120,878 & 2,550,209 \\
 Nigerian Fulfulde & \texttt{fuv\_Latn} & Atlantic-Congo & Latin && 919 & 4,281 & 264,234 \\
 West Central Oromo & \texttt{gaz\_Latn} & Afro-Asiatic & Latin && 3,399 & 9,071 & 1,640,693 \\
 Scottish Gaelic & \texttt{gla\_Latn} & Indo-European & Latin && 19,638 & 105,937 & 13,119,348 \\
 Irish & \texttt{gle\_Latn} & Indo-European & Latin && 60,303 & 267,562 & 45,341,371 \\
 Galician & \texttt{glg\_Latn} & Indo-European & Latin && 410,489 & 1,696,763 & 197,685,077 \\
 Guarani & \texttt{grn\_Latn} & Tupian & Latin && 207,800 & 1,038,296 & 48,610,979 \\
 Gujarati & \texttt{guj\_Gujr} & Indo-European & Gujarati && 21,916 & 87,805 & 3,202,096 \\
 Haitian Creole & \texttt{hat\_Latn} & Indo-European & Latin && 105,777 & 667,801 & 34,261,838 \\
 Hausa & \texttt{hau\_Latn} & Afro-Asiatic & Latin && 21,850 & 81,141 & 11,807,898 \\
 Hebrew & \texttt{heb\_Hebr} & Afro-Asiatic & Hebrew && 1,098,800 & 4,708,947 & 859,238,720 \\
 Hindi & \texttt{hin\_Deva} & Indo-European & Devanagari && 543,928 & 1,745,222 & 118,903,998 \\
 Chhattisgarhi & \texttt{hne\_Deva} & Indo-European & Devanagari && 832 & 3,908 & 205,345 \\
 Croatian & \texttt{hrv\_Latn} & Indo-European & Latin && 1,689,553 & 8,315,237 & 998,928,993 \\
 Hungarian & \texttt{hun\_Latn} & Uralic & Latin && 3,515,058 & 15,293,132 & 2,811,446,583 \\
 Armenian & \texttt{hye\_Armn} & Indo-European & Armenian && 336,285 & 1,126,920 & 199,883,484 \\
 Igbo & \texttt{ibo\_Latn} & Atlantic-Congo & Latin && 7,089 & 41,672 & 3,014,602 \\
 Ilocano & \texttt{ilo\_Latn} & Austronesian & Latin && 7,076 & 59,327 & 832,454 \\
 Indonesian & \texttt{ind\_Latn} & Austronesian & Latin && 6,644,918 & 16,237,247 & 2,895,956,979 \\
 Icelandic & \texttt{isl\_Latn} & Indo-European & Latin && 239,195 & 1,003,522 & 131,308,802 \\
 Italian & \texttt{ita\_Latn} & Indo-European & Latin && 12,812,932 & 47,011,085 & 8,144,757,759 \\
 Javanese & \texttt{jav\_Latn} & Austronesian & Latin && 18,192 & 100,952 & 15,206,708 \\
 Japanese & \texttt{jpn\_Jpan} & Japonic & Kanji && 14,154,575 & 23,435,549 & 8,539,956,266 \\
 Kabyle & \texttt{kab\_Latn} & Afro-Asiatic & Latin && 6,101 & 33,923 & 1,781,992 \\
 Kannada & \texttt{kan\_Knda} & Dravidian & Kannada && 9,373 & 33,147 & 1,206,651 \\
 Kashmiri & \texttt{kas\_Arab} & Indo-European & Arabic && 1,498 & 5,284 & 3,384,394 \\
 Georgian & \texttt{kat\_Geor} & Kartvelian & Georgian && 353,471 & 1,300,710 & 274,042,522 \\
 Kazakh & \texttt{kaz\_Cyrl} & Turkic & Cyrillic && 248,403 & 718,126 & 138,597,176 \\
 Halh Mongolian & \texttt{khk\_Cyrl} & Mongolic & Cyrillic && 123,789 & 505,098 & 83,628,495 \\
 Khmer & \texttt{khm\_Khmr} & Austroasiatic & Kher && 23,348 & 116,437 & 2,915,205 \\
 Kinyarwanda & \texttt{kin\_Latn} & Atlantic-Congo & Latin && 20,381 & 108,280 & 10,268,334 \\
 Kyrgyz & \texttt{kir\_Cyrl} & Uralic & Cyrillic && 51,221 & 194,092 & 33,981,180 \\
 Northern Kurdish & \texttt{kmr\_Latn} & Indo-European & Latin && 34,593 & 142,634 & 21,972,155 \\
 Korean & \texttt{kor\_Hang} & Koreanic & Hanja && 2,614,038 & 13,562,957 & 2,000,344,511 \\
 Lao & \texttt{lao\_Laoo} & Kra-Dai & Lao && 49,925 & 205,452 & 30,098,274 \\
 Ligurian & \texttt{lij\_Latn} & Indo-European & Latin && 3,581 & 26,740 & 1,046,463 \\
 Limburgish & \texttt{lim\_Latn} & Indo-European & Latin && 70,099 & 443,903 & 25,465,590 \\
 Lingala & \texttt{lin\_Latn} & Atlantic-Congo & Latin && 6,304 & 41,400 & 1,580,536 \\
 Lithuanian & \texttt{lit\_Latn} & Indo-European & Latin && 1,673,790 & 8,772,570 & 1,153,604,941 \\
 Lombard & \texttt{lmo\_Latn} & Indo-European & Latin && 14,053 & 61,359 & 6,270,646 \\
 Latgalian & \texttt{ltg\_Latn} & Indo-European & Latin && 5,174 & 21,062 & 2,903,043 \\
 Luxembourgish & \texttt{ltz\_Latn} & Indo-European & Latin && 27,946 & 142,470 & 13,925,521 \\
 Ganda & \texttt{lug\_Latn} & Afro-Asiatic & Latin && 1,475 & 4,118 & 688,308 \\
 Mizo & \texttt{lus\_Latn} & Sino-Tibetan & Latin && 7,009 & 22,630 & 4,106,536 \\
 Standard Latvian & \texttt{lvs\_Latn} & Indo-European & Latin && 857,757 & 3,937,940 & 578,441,751 \\
 Magahi & \texttt{mag\_Deva} & Indo-European & Devanagari && 290 & 1,088 & 94,031 \\
 Malayalam & \texttt{mal\_Mlym} & Dravidian & Malayalam && 11,203 & 44,417 & 1,420,906 \\
 Marathi & \texttt{mar\_Deva} & Indo-European & Devanagari && 43,720 & 142,001 & 6,164,176 \\
 Minangkabau & \texttt{min\_Latn} & Austronesian & Latin && 1,523 & 7,300 & 447,320 \\
 Macedonian & \texttt{mkd\_Cyrl} & Indo-European & Cyrillic && 539,149 & 1,841,846 & 304,592,615 \\
 Maltese & \texttt{mlt\_Latn} & Afro-Asiatic & Latin && 56,666 & 327,331 & 27,114,870 \\
 Maori & \texttt{mri\_Latn} & Austronesian & Latin && 20,840 & 114,680 & 24,524,962 \\
 Burmese & \texttt{mya\_Mymr} & Sino-Tibetan & Mon  && 6,575 & 36,661 & 406,016 \\
 Dutch & \texttt{nld\_Latn} & Indo-European & Latin && 16,890,074 & 64,609,055 & 9,493,533,101 \\
 Norwegian Nynorsk & \texttt{nno\_Latn} & Indo-European & Latin && 138,384 & 701,972 & 57,812,652 \\
 Norwegian Bokmål & \texttt{nob\_Latn} & Indo-European & Latin && 2,192,012 & 9,534,178 & 1,267,421,216 \\
 Nepali & \texttt{npi\_Deva} & Indo-European & Devanagari && 28,042 & 116,363 & 2,892,865 \\
 Nyanja & \texttt{nya\_Latn} & Atlantic-Congo & Latin && 11,749 & 65,324 & 8,513,823 \\
 Occitan & \texttt{oci\_Latn} & Indo-European & Latin && 61,681 & 323,632 & 21,029,975 \\
 Odia & \texttt{ory\_Orya} & Indo-European & Odia && 3,759 & 14,373 & 340,695 \\
 Pangasinan & \texttt{pag\_Latn} & Austronesian & Latin && 1,045 & 7,770 & 270,363 \\
 Eastern Panjabi & \texttt{pan\_Guru} & Indo-European & Gurmukhi && 10,857 & 44,440 & 1,821,511 \\
 Papiamento & \texttt{pap\_Latn} & Indo-European & Latin && 29,564 & 177,229 & 7,396,392 \\
 Southern Pashto & \texttt{pbt\_Arab} & Indo-European & Arabic && 31,854 & 107,563 & 27,623,486 \\
 Western Persian & \texttt{pes\_Arab} & Indo-European & Arabic && 6,995,368 & 24,998,370 & 6,061,794,870 \\
 Plateau Malgasy & \texttt{plt\_Latn} & Austronesian & Latin && 32,119 & 119,506 & 28,542,084 \\
 Polish & \texttt{pol\_Latn} & Indo-European & Latin && 14,492,239 & 60,362,860 & 10,994,239,010 \\
 Portuguese & \texttt{por\_Latn} & Indo-European & Latin && 8,033,406 & 26,058,040 & 4,639,089,792 \\
 Dari & \texttt{prs\_Arab} & Indo-European & Arabic && 421,097 & 2,101,038 & 399,037,437 \\
 Ayacucho Quechua & \texttt{quy\_Latn} & Quechuan & Latin && 1,248 & 10,038 & 322,112 \\
 Romanian & \texttt{ron\_Latn} & Indo-European & Latin && 5,131,444 & 17,790,793 & 3,484,865,185 \\
 Rundi & \texttt{run\_Latn} & Atlantic-Congo & Latin && 17,798 & 55,060 & 8,140,230 \\
 Russian & \texttt{rus\_Cyrl} & Indo-European & Cyrillic && 15,753,144 & 68,786,134 & 18,196,141,357 \\
 Sango & \texttt{sag\_Latn} & Atlantic-Congo & Latin && 724 & 4,564 & 181,876 \\
 Sicilian & \texttt{scn\_Latn} & Indo-European & Latin && 27,388 & 164,772 & 17,535,500 \\
 Sinhala & \texttt{sin\_Sinh} & Indo-European &  Sinhalese && 44,963 & 179,082 & 11,413,044 \\
 Slovak & \texttt{slk\_Latn} & Indo-European & Latin && 2,979,681 & 14,894,160 & 1,951,406,321 \\
 Slovenian & \texttt{slv\_Latn} & Indo-European & Latin && 1,456,026 & 7,106,291 & 928,101,642 \\
 Samoan & \texttt{smo\_Latn} & Austronesian& Latin && 11,024 & 62,358 & 11,672,900 \\
 Shona & \texttt{sna\_Latn} & Atlantic-Congo & Latin && 7,400 & 41,385 & 5,276,139 \\
 Sindhi & \texttt{snd\_Arab} & Indo-European & Arabic && 20,615 & 70,992 & 16,686,668 \\
 Somali & \texttt{som\_Latn} & Afro-Asiatic& Latin && 58,151 & 209,905 & 31,093,227 \\
 Southern Sotho & \texttt{sot\_Latn} & Atlantic-Congo & Latin && 7,474 & 41,714 & 5,876,842 \\
 Spanish & \texttt{spa\_Latn} & Indo-European & Latin && 22,218,630 & 76,372,709 & 13,882,047,139 \\
 Sardinian & \texttt{srd\_Latn} & Indo-European & Latin && 336,476 & 2,220,976 & 68,281,992 \\
 Serbian & \texttt{srp\_Cyrl} & Indo-European & Cyrillic && 593,332 & 2,251,042 & 394,477,097 \\
 Sundanese & \texttt{sun\_Latn} & Austronesian & Latin && 16,438 & 89,379 & 9,549,957 \\
 Swedish & \texttt{swe\_Latn} & Indo-European & Latin && 3,231,753 & 10,558,719 & 1,748,495,813 \\
 Swahili & \texttt{swh\_Latn} & Atlantic-Congo & Latin && 96,770 & 365,792 & 52,827,863 \\
 Silesian & \texttt{szl\_Latn} & Indo-European & Latin && 7,846 & 47,313 & 3,022,502 \\
 Tamil & \texttt{tam\_Taml} & Dravidian & Tamil && 30,202 & 149,837 & 4,234,345 \\
 Tatar & \texttt{tat\_Cyrl} & Turkic & Cyrillic && 34,489 & 133,014 & 22,255,423 \\
 Telugu & \texttt{tel\_Telu} & Dravidian & Telugu && 16,107 & 54,100 & 1,633,579 \\
 Tajik & \texttt{tgk\_Cyrl} & Turkic & Cyrillic && 119,383 & 395,470 & 87,519,228 \\
 Tagalog & \texttt{tgl\_Latn} & Austronesian & Latin && 140,922 & 628,210 & 95,285,900 \\
 Thai & \texttt{tha\_Thai} & Kra-Dai & Thai && 1,799,735 & 6,603,060 & 807,374,946 \\
 Tigrinya & \texttt{tir\_Ethi} & Afro-Asiatic & Ge‘ez && 2,622 & 8,601 & 1,699,272 \\
 Tok Pisin & \texttt{tpi\_Latn} & Indo-European & Latin && 785 & 5,888 & 97,298 \\
 Turkmen & \texttt{tuk\_Latn} & Turkic & Latin && 12,372 & 54,002 & 9,650,172 \\
 Turkish & \texttt{tur\_Latn} & Turkic & Latin && 4,448,111 & 12,304,912 & 2,356,627,784 \\
 Twi & \texttt{twi\_Latn} & Atlantic-Congo & Latin && 286 & 2,041 & 78,227 \\
 Uyghur & \texttt{uig\_Arab} & Turkic & Arabic && 10,614 & 41,367 & 6,602,690 \\
 Ukrainian & \texttt{ukr\_Cyrl} & Indo-European & Cyrillic && 2,689,369 & 10,842,572 & 1,909,330,669 \\
 Urdu & \texttt{urd\_Arab} & Indo-European & Arabic && 403,245 & 1,224,175 & 236,356,788 \\
 Northern Uzbek & \texttt{uzn\_Latn} & Turkic & Latin && 113,772 & 581,861 & 81,808,833 \\
 Venetian & \texttt{vec\_Latn} & Indo-European & Latin && 122,390 & 763,029 & 24,081,966 \\
 Vietnamese & \texttt{vie\_Latn} & Viet-Muong & Latin && 12,296,989 & 46,339,341 & 11,462,111,787 \\
 Wolof & \texttt{wol\_Latn} & Atlantic-Congo & Latin && 2,152 & 9,351 & 367,848 \\
 Xhosa & \texttt{xho\_Latn} & Atlantic-Congo & Latin && 13,620 & 80,748 & 14,566,904 \\
 Eastern Yiddish & \texttt{ydd\_Hebr} & Indo-European & Hebrew && 12,275 & 56,421 & 17,078,751 \\
 Yoruba & \texttt{yor\_Latn} & Atlantic-Congo & Latin && 10,148 & 49,474 & 8,346,193 \\
 Yue Chinese & \texttt{yue\_Hant} & Sino-Tibetan & Hant && 28,478 & 172,592 & 21,579,579 \\
 Chinese (Simplified) & \texttt{zho\_Hans} & Sino-Tibetan & Hanzi && 8,326,440 & 29,575,591 & 5,199,137,981 \\
 Chinese (Traditional) & \texttt{zho\_Hant} & Sino-Tibetan & Hant && 3,796,336 & 15,514,804 & 2,617,463,485 \\
 Standard Malay & \texttt{zsm\_Latn} & Austronesian & Latin && 864,831 & 3,651,754 & 384,708,004 \\
 Zulu & \texttt{zul\_Latn} & Atlantic-Congo & Latin && 13,089 & 73,167 & 9,654,461
    \label{tab:languages-stats}
\end{longtable}}
\twocolumn

\begin{table*}
    \centering
    \resizebox{\textwidth}{!}{\begin{tabular}{lr} \toprule

    Relevant words describing the topics & Topic representation (in \%) \\ \midrule
       recipe, sauce, cheese, recipes, chicken, add, delicious, food, cook, minutes & 7.17 \\
game, games, players, gaming, play, gameplay, pc, playing, player, playstation & 2.1 \\
dog, dogs, cat, pet, cats, pets, puppy, breed, puppies, animal & 1.94 \\
god, jesus, church, christ, faith, lord, bible, him, christian, holy & 1.82 \\
shirt, jacket, jeans, cotton, wear, shirts, fit, men, size, waist & 1.68 \\
estate, property, mortgage, real, home, house, buyers, market, properties, homes & 1.6 \\
art, artist, artists, painting, gallery, paintings, works, arts, his, exhibition & 1.59 \\
card, cards, love, gift, cute, christmas, fun, stamp, halloween, easter & 1.57 \\
covid, 19, vaccine, coronavirus, virus, cases, health, vaccinated, vaccination, vaccines & 1.28 \\
album, band, music, song, songs, rock, release, albums, his & 1.27 \\
data, cloud, server, software, aws, application, business, management, cluster, configuration & 1.21 \\
shipping, item, delivery, ankle, brace, items, order, days & 1.14 \\
diamond, jewelry, ring, carat, gold, rings, earrings, silver, necklace & 1.13 \\
books, book, read, she, reading, author, novel, story & 1.13 \\
dress, fashion, dresses, wear, style, skirt, wedding, outfit, love & 1.12 \\
bedroom, apartments, room, property, apartment, kitchen, floor, home, spacious & 1.03 \\
trump, president, election, biden, republican, democrats, senate, republicans, court & 1.0 \\
bitcoin, crypto, cryptocurrency, blockchain, ethereum, tokens, token, price, cryptocurrencies, nft & 0.94 \\
life, tarot, yourself, love, feel, myself, mind & 0.93 \\
sound, hearing, audio, speakers, noise, headphones, bluetooth, speaker, microphone, wireless & 0.9 \\
trail, lake, park, mountain, hike, trails, hiking, canyon, river, yosemite & 0.85 \\
cookies, website, cookie, settings, resume, preferences, disable, browser, user, information & 0.85 \\
police, said, arrested, man, officers, crime, county & 0.84 \\
skin, acne, skincare, sunscreen, face, treatment, facial, pores, oil, hyamax & 0.83 \\
casino, gambling, games, slot, casinos, online, bonus, poker, slots, players & 0.82 \\
lighting, light, lights, led, lamp, bulb, lamps, bulbs, wall, fixtures & 0.76 \\
league, club, chelsea, season, his, football, cup, players, player & 0.76 \\
wedding, weddings, venue, couples, dj, guests, day, reception, ceremony, bride & 0.76 \\
dental, teeth, tooth, dentist, smile, dentistry, whitening, oral, implant, dentists & 0.72 \\
cbd, cannabis, marijuana, hemp, thc, gummies, cannabinoids, medical, oil, recreational & 0.69 \\
shoes, shoe, boots, socks, nike, sneaker, heel, foot, running, boxing & 0.68 \\
stocks, inflation, stock, market, investment, gold, investors, fund, trading, funds & 0.67 \\
car, cars, engine, rear, porsche, toyota, mercedes, electric, suv & 0.65 \\
fitness, workout, exercise, gym, trainer, exercises, training, strength, body, workouts & 0.63 \\
divorce, attorney, law, lawyer, legal, court, lawyers, attorneys, injury, case & 0.59 \\
steel, nailer, arm, drill, machine, wrench, tool, saw, palm & 0.59 \\
security, cyber, ransomware, cybersecurity, attacks, malware, phishing, attack, data, hackers & 0.59 \\
furniture, chair, chairs, sofa, table, design, style, dining, comfort, leather & 0.5 \\
her, she, actress, star, kristina, instagram, lorena, grikaite, kardashian & 0.49 \\
testosterone, vitamin, supplements, weight, blood, body, supplement, levels, calcium, muscle & 0.49 \\
bag, bags, backpack, stethoscope, pockets, strap, tote, zipper, purse, leather & 0.48 \\
energy, carbon, wind, renewable, emissions, gas, coal, power, electricity, climate & 0.47 \\
ball, tennis, player, sports, sport, players, backhand, coach, athletes, drills & 0.47 \\
phone, samsung, smartphone, camera, galaxy, oneplus, realme, battery, schematic, android & 0.47 \\
watches, replica, watch, rolex, dial, clock, patek, swiss, fake, wholesale & 0.47 \\
fundraising, volunteer, donors, volunteers, donor, charity, charities, community, volunteering, imdsa & 0.46 \\
parking, traffic, pedestrian, rail, transport, transportation, street, mobility, public, city & 0.45 \\
race, f1, racing, lap, formula, drivers, season, car, championship, driver & 0.44 \\
pain, chiropractic, joint, ankle, spinal, muscles, arthritis, spine, symptoms, joints & 0.41 \\
wine, wines, winery, bordeaux, vineyard, tasting, sauvignon, grapes, vineyards, palate & 0.41 \\ \bottomrule
    \end{tabular}}
    \caption{BERTopic modeling on the \textbf{English} subset of mOSCAR.}
    \label{topic-mode-en}
\end{table*}

\begin{table*}
    \centering
    \resizebox{\textwidth}{!}{\begin{tabular}{lr} \toprule

    Relevant words describing the topics & Topic representation (in \%) \\ \midrule
    
    Kosovo, Serbia, European, taken, Kurti, Serbian, Serbia & 4.65 \\

Ukraine, Russian, Russia, Putin, war, Ukrainian, military & 4.44 \\

Covid, 19, vaccine, cases, health, patients, coronavirus & 3.72 \\

Luizi, Kiara, Big, Brother, VIP, Olta, stuck & 3.17 \\

apartment, rent, floor, m2, lease, housing, located, area, sale & 2.51 \\

letter, language, book, literary, poetry, book, Albanian, writer & 1.81 \\

add, spoon, eggs, recipe, sugar, pour, oven, oil, flour & 1.46 \\

photo, she, Instagram, hers, bikini, Tarja, model, follows, dress & 1.23 \\

Allah, greeting, Quran, prophet, interpretation, exegesis, Islam, prophets & 1.19 \\

court, property, court, trials, KPK, reevaluation, appeal, judges, declaration, prosecution & 1.16 \\

horoscope, sign, signs, you, stars, zodiac & 0.99 \\

music, festival, stage, KultPlus, artist, culture, musical & 0.98 \\

prison, court, criminal, imprisonment, pretrial detention, measure, sentenced, crime & 0.85 \\

Inter, Inter Milan, Milan, AC Milan, Gazzetta, Pioli & 0.82 \\

Rama, Edi, prime minister, opposition, PD, McGonigal, Basha, justice, PS & 0.82 \\

Trump, Biden, Donald, president, Republican, president’s, Joe, American, White, president & 0.81 \\

accident, occurred, consequence, road, car, injured, type, crashed & 0.78 \\

Berisha, PD, Sali, Berisha, Rama, opposition, elections, Rama, Edi, party’s & 0.74 \\

exhibition, art, museum, exhibition, culture, artist, KultPlus & 0.72 \\

knife, gold, weapon, police, killed, incident, fired, event, victim & 0.7 \\

water, river, residents, floods, rain, drinking & 0.7 \\

women, gender, violence, LGBT, friend, violence, family & 0.68 \\

migrants, asylum, refugees, permits, Albanian, passport, exchange, asylum seeker, 000, passport & 0.63 \\

students, UET, academic, university, university’s, Erasmus, universities, faculty, study, UBT & 0.63 \\

tourists, tourism, tourist, travel, foreign, destination, Albanian, visitors & 0.60 \\

doctors, patients, health, clinic, hour, sore, clinic, Dr & 0.57 \\

firefighters, fire, flame, fires, extinguishing, hotspots & 0.56 \\

Ronaldo, Cristiano, CR7, Portuguese, United, Manchester, Messi & 0.56 \\

music, musical, Adriano, Celentano, musical, Tari, top, titled & 0.56 \\

weather, temperatures, forecast, cloudy, rain, snow, degrees & 0.56 \\

Mercedes, car, Audi, sale, BMW, Aston, vehicles, Benz, sold & 0.52 \\

email, site, comment, browser, save, address, fields, marked, required & 0.5 \\

archaeologists, archaeological, Roman, ancient, city, archaeological, restored, Durres, century & 0.49 \\

NASA, Earth, space, science, planet, boundary, solar, Apollo, planets & 0.47 \\

Albanian, November, flag, nations, independence, year, day, Vasfije & 0.46 \\

oil, liter, price, fuel, lek prices, gasoline, petrol, barrel & 0.45 \\

PD, Democratic, party’s, party, PS, elections, vote, party, Socialist, Democrat & 0.44 \\

city, located, park, rock, water, natural & 0.44 \\

Turkey, earthquake, Syria, magnitude, collapsed & 0.43 \\

airline, plane, passengers, flight & 0.41 \\

road, buses, urban, bike, works, city’s, urban, municipality, Veliaj, transportation & 0.4 \\

iPhone, Apple, Pro, Max, iOS, iPad, 15, apps, 14, dollars & 0.4 \\

north, North Macedonia, Bulgarian, Bulgaria, Macedonian, Macedonia, Macedonians & 0.4 \\
        \bottomrule
    \end{tabular}}
    \caption{BERTopic modeling on the \textbf{Tosk Albanian} subset of mOSCAR.}
    \label{topic-mode-als}
\end{table*}

\begin{table*}
    \centering
    \resizebox{\textwidth}{!}{\begin{tabular}{lr} \toprule

    Relevant words describing the topics & Topic representation (in \%) \\ \midrule

Match, league, Madrid, Barcelona, match, football, Real & 9.86 \\

Allah, Sheikh, Mohammed, occupation & 4.24 \\

Allah, cleaning, homes, jurisprudential, buying, Kuwait, furniture & 2.78 \\

Game, games, download, casino, Android, play, computer & 2.5 \\

Sarai, occupation, Jerusalem, Palestinian, Gaza, next, Palestine & 2.08 \\

Corona, health, virus, infection, Covid, 19, vaccine & 1.98 \\

Ukraine, Russia, Russian, Putin, Moscow, Kiev, forces, Ukrainian & 1.86 \\

Fridge, LED, electric, watts, stainless steel, product, resistant, corrosion & 1.66 \\

Movie, film, TV series, actress, actor, series, cinematic, ceremony, cinema & 1.54 \\

Hospital, medical, health, medical, care, doctors, diseases, doctor & 1.47 \\

Car, cars, Toyota, Hyundai, electric, Mercedes, Ford, Kia & 1.44 \\

Capital, Daraya, compound, real estate, mall, meters, administrative, units, for sale & 1.4 \\

Data, workers, feasibility, project, work, study, customers, employees & 1.25 \\

Syrian, Syria, ISIS, Aleppo, forces & 1.01 \\

Saudi, national, Saudi Arabia, Kingdom, Arabia, Muhammad & 0.98 \\

Aviation, airport, flight, airplane, aircraft, air, airlines & 0.91 \\

Women, women’s, violence, rights, society, feminist, affected, victims, origin & 0.83 \\

iPhone, iPhone, Apple, iOS, phone, iPad, apps & 0.83 \\

Space, NASA, Mars, moon, planet, planets, scientists, telescope & 0.81 \\

Iran, Tehran, nuclear, agreement, Washington, United States, American & 0.79 \\

Tourism, hotel, island, Maldives, beach, Malaysia, shores & 0.71 \\

Skin, cream, face, skin, oil, care, your skin, treatment, oily & 0.65 \\

Chocolate, spoon, meat, recipes, cake, butter, method, recipe, bowl & 0.62 \\

College, engineering, science, education, university, sciences, department, school & 0.6 \\

Temperature, weather, degrees, forecast, airport, temperature, west, regions & 0.6 \\

Police, suspect, arrest, drugs, crime, general, city, prosecutor & 0.59 \\

Spine, column, knee, blood, body, fever, pain, muscle & 0.59 \\

Istanbul, Turkey, Turkish, apartments, for sale, real estate, nationality, Şişli, property & 0.59 \\

Pregnancy, fetus, womb, birth, monthly, cycle, doctor, progesterone, symptoms & 0.57 \\

Sudan, Ethiopia, Tigray, Ethiopian, Sudanese, Ethiopian, Renaissance, army, Sudanese & 0.55 \\

Germany, Sweden, Europe, asylum, Merkel, immigration, refugee & 0.54 \\

Insects, pest control, ants, pests, spray, pesticides, cockroaches, company, white & 0.53 \\

University, Baniyas, scholarships, students, universities, study & 0.51 \\

Gold, price, carat, today, pound, grams, dealings & 0.5 \\

Medicine, drug, prescription, dosage, side effects, doctor, treatment, illness & 0.5 \\

Rooms, bedrooms, decorations, modern, wallpaper, furniture, sleeping & 0.49 \\

Hotels, activities, trips, close, hotel, tours & 0.45 \\

Samsung, Galaxy, phone, specifications, S23, 5G, phones & 0.44 \\

Fire, accident, firefighters, accident, Greece, outbreak, injured, forest & 0.42 \\

Trump, Donald, Biden, elections, American, former, White House & 0.41 \\

Technician, painter, Kuwait, repair, installation, AC, satellite, changing, provided & 0.4 \\

Hair, hair loss, shampoo, scalp, grow, hair, hair treatment & 0.39 \\

Education, secondary, studies, students, exams, schools, cumulative, average & 0.39 \\

Elections, voting, election, Myanmar, council, parties, electoral & 0.39 \\

Buyer, seller, shipment, equipment, Alibaba, help, charger, suppliers, machine, tank & 0.39 \\

India, ambassador, cooperation, Libya, ministers, Modi & 0.39 \\

        \bottomrule
    \end{tabular}}
    \caption{BERTopic modeling on the \textbf{Arabic} subset of mOSCAR.}
    \label{topic-mode-ar}
\end{table*}

\begin{table*}
    \centering
    \resizebox{\textwidth}{!}{\begin{tabular}{lr} \toprule

    Relevant words describing the topics & Topic representation (in \%) \\ \midrule

team, match, child, event, olympic, goal, championship, win, champion & 10.84 \\

school, students, education, children, second, student & 4.11 \\

elections, syriza, mitsotakis, government, ND, response, prime minister, parliament, no & 2.75 \\

order, shipment, product, payment, purchase, cash on delivery, receipt, donation, you, shipping fees & 2.15 \\

god, god's, church, holy, christ, priest, prayer, divine & 1.8 \\

gold, necklace, earrings, ring, bracelet, jewelry, silver, steel & 1.79 \\

book, writer, history, published, read, literature, write, novel & 1.64 \\

game, play, children, two, ball, games & 1.36 \\

ukraine, russia, russian, force, usa, war, russia & 1.21 \\

shoes, t-shirt, adidas, nike, sneakers, men, athletic, running & 1.18 \\

song, music, album, single, music, two, rock & 1.16 \\

skin, face, cream, body, oil, oily, skin care & 1.15 \\

traffic, vehicle, road, lane, road, vehicles, street & 1.11 \\

mattress, sofa, pillow, bed, chair, furniture, armchair, bedroom, color & 1.04 \\

recipe, chicken, recipe, juicy, add, chicken, mix, season & 1.02 \\

christmas, birth, christmas, december, gift, festive, tree, holiday & 0.98 \\

turkey, erdogan, turkish, greece, home, democracy, istanbul & 0.93 \\

moon, planet, nasa, sky, galaxy, earth, telescope, satellite, lunar, space & 0.83 \\

tourism, tourist, greece, destination, trips, hotel & 0.81 \\

mayor, march, meeting, municipal, council, committee, session, members & 0.81 \\

ships, boats, port, cruise, shipping, sea, cruise ship, peiraeus & 0.78 \\

fire, fire brigade, wildfire, fire extinguishers, area, blaze, fire vehicles, firemen & 0.78 \\

beach, island, water, village, white, place & 0.77 \\

art, painting, works, artist, museum, exhibition, artists & 0.76 \\

car, bmw, model, new, nissan, audi, mobile, video & 0.75 \\

vaccine, vaccination, covid, doses, health, coronavirus, disease & 0.74 \\

led, lighting, lamp, simple, lights, white, lighting, hanging, ceiling, lamps & 0.72 \\

you, yourself, feeling, life, thoughts, live, feel & 0.68 \\

cotton, cotton, fabric, design, cotton & 0.67 \\

purse, wallet, leather, case, bag, items & 0.65 \\

exercise, gym, workout, practice, body, strength, pilates, muscle & 0.64 \\

weather, rainy, temperature, clouds, cold, storms, celsius, wind & 0.63 \\

company, product, sunlight, quality, service, construction, production, service & 0.62 \\

plants, seeds, plant, grow, flowers, neem, soil, wild & 0.62 \\

europe, euro, greek, eu, policy, government, eurozone, economy & 0.61 \\

wine, wines, vineyard, winery, variety, grapes, alcohol & 0.57 \\

pump, burner, gas, water, temperature, heating, air, water heater & 0.55 \\

headphones, bluetooth, speakers, sound, usb, wireless, audio, sound & 0.54 \\

movie, series, netflix, characters, director, episode, season, naruto & 0.54 \\

apartment, for sale, bedroom, location, new, living, for rent, kitchen & 0.49 \\

restaurant, kitchen, chef, dish, food, menu, restaurant, tavern & 0.49 \\

diet, weight, food, loss, calories, pounds, healthy & 0.48 \\

history, news, breakfast, syriza, greek, government, politics, news & 0.47 \\

video, tv, sony, iptv, dvd, photo, movie, camera & 0.47 \\

dog, dogs, pet, animal, dog, puppies, leash, friend, pets & 0.46 \\

inflation, rates, euro, increase, prices, interest, market, economy & 0.42 \\

        \bottomrule
    \end{tabular}}
    \caption{BERTopic modeling on the \textbf{Greek} subset of mOSCAR.}
    \label{topic-mode-el}
\end{table*}

\begin{table*}
    \centering
    \resizebox{\textwidth}{!}{\begin{tabular}{lr} \toprule

    Relevant words describing the topics & Topic representation (in \%) \\ \midrule

loan, bank, paytm, sbi, upi, credit, personal, sunal, atm, account & 3.61 \\

electric, tata, mahindra, hyundai, maruti, bike, suv & 3.16 \\

recipe, make, paneer, chicken, best, ingredients, potato & 2.45 \\

recruitment, notification, availability, 2023, vacancy, age, position, fill & 1.39 \\

whatsapp, sap, fm, gb, chat, gbwhatsapp, message, group, feature & 1.28 \\

movie, khan, film, jawan, trailer, release, series, jaari, cast, collection & 1.25 \\

bitcoin, cryptocurrency, token, crypto, currency, ethereum, coin, blockchain, binance, dogecoin & 1.25 \\

samsung, 5g, oneplus, realme, redmi, nokia, oppo, vivo, galaxy, xiaomi & 1.19 \\

vs, ind, aus, odi, one-day, wicket, match, runs, australia, cup & 1.1 \\

stock, market, trading, intraday, share, ratio, demat, year, buy & 0.88 \\

wish, life, dog, equity, confidence, morning, suprabhat, habit, flirting, self & 0.86 \\

business, ideas, idea, manufacturing, making, small, entrepreneur, start, recycling, people & 0.85 \\

aadhar, aadhaar, uidai, half, card, otp, update, number, pan, bar & 0.72 \\

designs, design, taarak, mehta, alto, jewellery, mangalsutra, galas, blouse, earrings & 0.72 \\

rlalibaba, equipment, mash, ravi, gati, jak, shakash, ens, quot, woodworking & 0.69 \\

iphone, apple, aif, apple, 14, ipad, 15, ios, xs, mac & 0.67 \\

biography, chay, kanika, mann, one, alia, bhatt, tulsidas, family, age & 0.63 \\

reply, cancel, leave, rac, berth, approval, seat, resignation, vishesh, vyakti & 0.6 \\

movies, movie, download, hollywood, tamil, dubbed, telugu, hd & 0.57 \\

shayari, attitude, love, dil, sad, status, romantic, nahin, mohabbat, girlfriend & 0.56 \\

ganga, festival, dussehra, jayanti, ekadashi, sav, janmashtami, bihu, raksha, dashera & 0.55 \\

computer, processor, tar, virus, system, software, year, input, output & 0.54 \\

yatan, thal, places, tourist, garan, yatak, kuchinda, jaipur, tourism, barot & 0.54 \\

course, bba, llb, pharma, bsc, ba, bachelor, mba, ed, entrance & 0.52 \\

ration, vein, dhaman, artery, card, shan, list, epds, fcs, nfsa & 0.51 \\

jio, phone, tune, caller, recharge, sim, plan, call, reliance, jiotv & 0.5 \\

youtube, channel, video, youtuber, sponsorship, shorts, subscriber, videos, views & 0.5 \\

train, railway, railways, trains, station, vay & 0.46 \\

freelancing, money, kam, earn, paise, online, kamaye, fiverr, earning, freelancer & 0.45 \\

weight, creatine, loss, pathri, at, diet, ghat, patal & 0.44 \\

photo, lightroom, background, image, photoshop, edit, presets, kag, preset, jpeg & 0.43 \\

shram, card, eshram, ram, check, payment, uan, shramik, otp, gov & 0.42 \\

election, nasbah, assembly, bjp, president, elections, powers, veto, gujarat & 0.4 \\

haven, matching, suppliers, supplier, verified, let, found, alibaba, equipment & 0.39 \\

stotram, lord, shiva, shani, dev, ganesha, gan, bhagav, kath & 0.39 \\

kisan, pm, beneficiary, farmer, pm kisan, corner, kyc, status, nidhi, samman & 0.38 \\

\bottomrule
    \end{tabular}}
    \caption{BERTopic modeling on the \textbf{Hindi} subset of mOSCAR.}
    \label{topic-mode-hi}
\end{table*}

\subsection{Heuristics to increase the quality of documents} \label{sec:heuristics}

We use a set of heuristics to improve the quality of the documents by discarding some text nodes. We first consider text nodes to be written in Latin scripts if more than 50\% of the characters are Latin. In detail, we discard the text node if:
\begin{enumerate}
    \item It is empty.
    \item It contains fewer than 5 bytes for Latin scripts and fewer than 15 bytes for non-Latin scripts.
    \item More than 30\% of the characters are digits.
    \item It contains more than one date.
    \item It contains the sequence ``lorem ipsum''.
    \item The ratio of non-alphabetic characters is superior to 0.33.
    \item The symbols `\{' or '`\}' are in the text.
    \item The symbols `$\geq$', `$\leq$', `>' or `<' are more than 2 times in the text.
    \item ``Follow us'', ``javascript'', ``copyright'' or ``©'' are in the text.
    \item The ratio of capitalized letters is superior to 0.2.
    \item The text exactly matches with ``comment'', ``facebook'', ``instagram'', ``twitter'', ``rss'', ``newsletter'', ``share'' or ``follow us''.
    \item A character is more than 33\% of the total number of characters in the string.
\end{enumerate}

We then also apply some filters to clean the text as much as possible:
\begin{enumerate}
    \item Remove URLs from all documents.
    \item Normalize consecutive special characters (`\textbackslash t', `\textbackslash n', `\#', `/', `\$', `)', `(', `[', `]', `!', `?', `\%', `<', `>') to keep only one.
\end{enumerate}

Following previous steps, we keep the text node if it is superior to 5 bytes and we keep the final document if it is superior to 100 bytes.

\subsection{Text-Image similarity and DOM Tree} \label{sec:txt-img-sim-analysis}

\begin{figure*}[htbp]
    \centering
    \begin{subfigure}{0.4\textwidth}
        \centering
        \includegraphics[width=.9\textwidth]{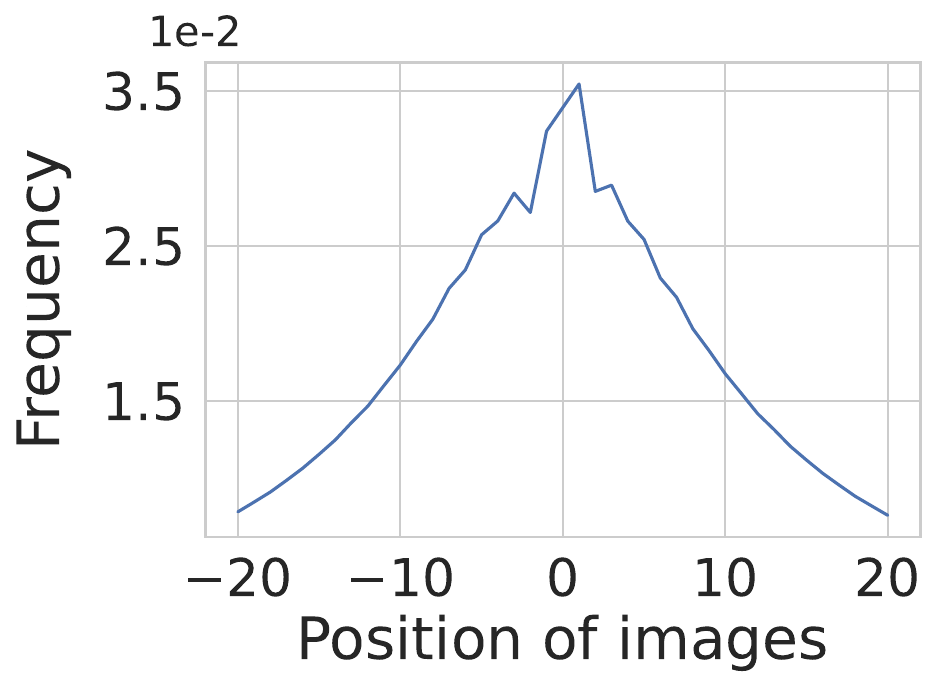}
        \caption{Relative position in the document of relevant text nodes with respect to images.}
        \label{fig:relative-node-pos-img}
    \end{subfigure}\hspace{5mm}
    \begin{subfigure}{0.4\textwidth}
        \centering
        \includegraphics[width=.9\textwidth]{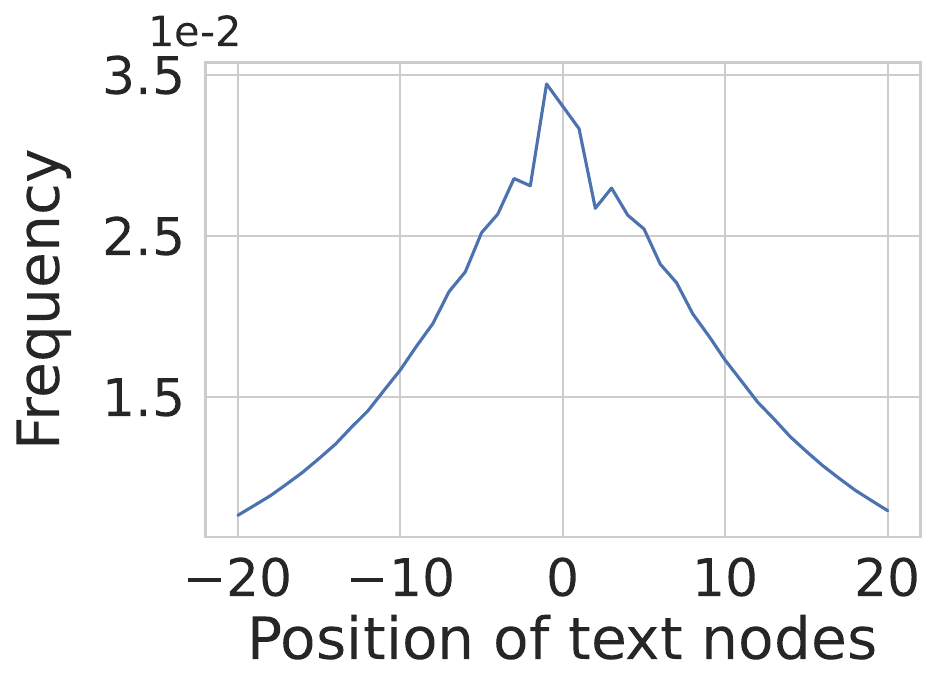}
        \caption{Relative position in the document of relevant images with respect to text nodes.}
        \label{fig:relative-node-pos-text}
    \end{subfigure}
    \caption{Relative positions of most relevant images and text nodes with respect to the other modality.}
    \label{fig:relative-node-pos}
\end{figure*}

As we rely on the DOM Tree to build the documents and the order of appearance of the nodes could differ from HTML rendering, we attempt to assess to what extent it is a relevant way of constructing a multimodal document. To do so, we rely on the results of the text-image joint filtering step where we compute the ranks of relevant text nodes (resp images) for each image. We plot the distribution of the closest most relevant node for each modality in \Cref{fig:relative-node-pos-img,fig:relative-node-pos-text}. We notice that the most relevant node to either a text node or an image is their closest node in the DOM tree. The cumulative distribution function of the distribution of the closest node reaches 25\% for nodes positioned between -5 and 5, which confirms the relevance of using the DOM tree to represent a document.

\subsection{Implementation details} \label{sec:impl-details}

\subsubsection{Text deduplication parameters} \label{sec:text-dedup-lsh}
Following previous work, we near-deduplicate documents using MinHashLSH. We first vectorize the documents using HashingVectorizer from scikit-learn with 2,097,152 features computed on 4-grams and 5-grams within word boundaries. We then compute MinHashes from those vectors with 256 permutations and we finally run Locality Sensitive Hashing with a threshold Jaccard Similarity of 0.8 for finding near-duplicates.

\subsubsection{Removing Personal Identifiable Information} \label{ss:pii}

We used regular expressions to detect and remove PII in documents. More precisely, we used: \\

\textbf{email address}:

\verb/^[\w\.]+@[\w-]+\.[\w-]{2,4}$/ \\

\textbf{phone number}:

\verb/^\+?\d{1,3}?[-.\s]?\(?\d{1,4}?\)?/

\verb/[-.\s]?\d{1,4}[-.\s]?\d{1,4}/
  
\verb/[-.\s]?\d{1,9}$/ \\

\textbf{credit card number}:

\verb/^(?:4[0-9]{12}(?:[0-9]{3})?|5[1-5]/

\verb/[0-9]{14}|3[47][0-9]{13}|3(?:0[0-5]|/

\verb/[68][0-9])[0-9]{11}|6(?:011|5[0-9]/

\verb/{2})[0-9]{12}|(?:2131|1800|35/

\verb/\d{3})\d{11})$/ \\

\textbf{IP address}: 

\verb/^(?:25[0-5]|2[0-4][0-9]|1[0-9]{2}|/

\verb/[1-9][0-9]|\d)\.(?:25[0-5]|2[0-4]/

\verb/[0-9]|1[0-9]{2}|[1-9][0-9]|\d)\./

\verb/(?:25[0-5]|2[0-4][0-9]|1[0-9]{2}|/

\verb/[1-9][0-9]|\d)\.(?:25[0-5]|2[0-4]/

\verb/[0-9]|1[0-9]{2}|[1-9][0-9]|\d)$/ \\

\textbf{passport number}: \verb/^[A-Z0-9]{6,15}$/ \\

For images, we detect faces in the images and distribute the bounding boxes coordinates. More precisely, all the images are resized to have a maximum of width and height of 256, keeping aspect ratio. The bounding boxes coordinates are therefore computed given this image size but can be extrapolated if images are downloaded in a higher resolution. 

\subsubsection{Training implementation details}
We train multilingual OpenFlamingo on mOSCAR and multilingual text-image pairs. We use a batch of size 64 for mOSCAR and 128 for captioning data, limiting the number of tokens to 256 for mOSCAR and 32 for captioning data. Similarly to Flamingo and OpenFlamingo, text tokens can only attend to the previous image in the sequence. To increase diversity in the training batch, we randomly reject 2/3 of the documents if they contain only one image. We limit the maximum number of images in a sequence to 8. We randomly sample 8 languages per batch and upsample low-resource languages. We train multilingual OpenFlamingo on 43 languages covering all the languages of the benchmarks we evaluate the models on (see Section~\ref{sec:eval-details}).

We use Gemma-2B as the underlying language model behind multilingual OpenFlamingo and CLIP ViT-L-14 as the image encoder. We add a cross-attention layer after each decoder layer. Following OpenFlamingo, we add the two special tokens \texttt{<image>} and\texttt{<|endofchunk|>}, whose embeddings were trained. Only the Perceiver Resampler, cross-attention layers and these two embeddings were trained; everything else remained frozen. During training, we apply a factor of 0.2 for the captioning data loss function.

We train the model using the Adam optimizer and a maximum learning rate of 1e-4. We use a constant learning rate scheduler with 1875 warm-up steps. We use 4 accumulation gradient steps to have an effective batch of size 256 for mOSCAR and 512 for captioning data. We train the model on 50M documents and 100M image-text pairs on 8 Nvidia A100 for 170h.

\subsubsection{Evaluation details} \label{sec:eval-details}

\begin{table*}[!htbp]
    \centering\small
    \resizebox{.9\textwidth}{!}{\begin{tabular}{llrl} \toprule
         & Metric & \#examples &  Languages \\ \midrule

       xFlickr\&CO & CideR & 2,000 & Chinese, English, German, Indonesian, Japanese, Russian, Spanish, Turkish \\
        
        & & & \\
       \multirow{4}{*}{XM3600} & \multirow{4}{*}{CideR} & \multirow{4}{*}{3,600} & Arabic, Czech, Danish, German, Greek, English, Spanish, Farsi, \\

       & & &  Finnish, French, Hebrew, Hindi, Croatian, Hungarian, Indonesian, Italian, \\

       & & & Japanese, Korean, Dutch, Norwegian, Poland, Portuguese, Romanian, \\

       & & & Russian, Swedish, Telugu, Thai, Turkish, Ukrainian, Vietnamese, Chinese \\

 & & & \\
 
       xGQA & Accuracy & 9,666 & Bengali, German, English, Indonesian, Korean, Portuguese, Russian, Chinese \\

        & & & \\

        MaXM & Accuracy & $\sim$ 170 & English, French, Hindi, Hebrew, Romanian, Thai, Chinese \\

        & & & \\

        MaRVL & Accuracy & $\sim$ 1,150 & Indonesian, Swahili, Tamil, Turkish, Chinese \\

        & & & \\

        XVNLI & Accuracy & 1,164 & English, Arabic, Spanish, French, Russian \\

        & & & \\

        Multi30k & BLEU & 1,000 & French, German, Czech \\

        & & & \\
        
       CoMMuTE  & Accuracy & 310 & Czech, French, German \\ \bottomrule
    \end{tabular}}
    \caption{Overview of the benchmarks used to evaluate our multilingual OpenFlamingo.}
    \label{tab:stats-benchmarks}
\end{table*}

We evaluate on a set of eight benchmarks: xFlickr\&CO, XM3600, xGQA, MaXM, MaRVL, XVNLI, Multi30k (Test2016 subset) and CoMMuTE; covering 5 different tasks and 43 languages. Details about the languages, the number of examples and the metric used can be found in Table~\ref{tab:stats-benchmarks}. We used the \textit{translate-test}\footnote{Benchmark automatically translated into English.} samples provided by the authors of the benchmarks if available. No translate test samples were provided for MaXM, so we translated the test set using the NLLB-600M distilled model. As no training set was available for MaXM, we use the few-shot examples from xGQA. Since we use Stanza tokenizers, we could not evaluate on all languages from XM3600 as 3 of them were not available. Filipino was also not into the list of mOSCAR languages, so we skip this language during evaluation. The CoMMuTE evaluation set involves choosing between two  different translations of a same source text (one correct and one incorrect depending on an image provided to disambiguate the text). We use the lowest perplexity between the two translations as the model's prediction. We also use Multi30k training set as few-shot examples.

\paragraph{Prompting}
Following previous works, the zero-shot setting is composed of two few-shot examples without providing the images. The prompts we use for the different tasks are as follows:\footnote{We show the prompts we used with one context example.}

For \underline{captioning} tasks, we use the prompt:

``\verb/<image>Output:[Caption]<|endofchunk|>/

\verb/<image>Output:/'',

where \texttt{[Caption]} is replaced by the caption. \\

For \underline{visual question answering} tasks, we use the prompt:

``\verb/<image>Question: [Question]/

\verb/Short Answer: [Answer] <|endofchunk|>/

\verb/<image>Question: [Question]/

\verb/Short Answer:/'',

where \texttt{[Question]} and \texttt{[Answer]} are replaced by the questions and the answer respectively. \\

For \underline{multimodal machine translation} tasks, we use the prompt: 

``\verb/<image>Sentence:`[Caption]'./

\verb/Translation: [Translation] <|endofchunk|>/

\verb/<image>Sentence:`[Caption]'/

\verb/Translation:/'',

where \texttt{[Caption]} is replaced by the sentences to translate and \texttt{[Translation]} is replaced by its translation. \\

For \underline{MaRVL}, we use the prompt: 

``\texttt{<image> `[Statement]'. True of False? [Answer]<|endofchunk|><image> `[Statement]'. True of False?}'',

where \texttt{[Statement]} is replaced by the statement and \texttt{[Answer]} by the answer. We also concatenate the left and right image into a single image. \\

For \underline{XVNLI}, we use the prompt: 

``\verb/<image> `[Statement1]' - `[Statement2]'./

\verb/entailment, neutral or contradiction?/ 

\verb/Output: [Answer]<|endofchunk|>/

\verb/<image> `[Statement1]' - `[Statement2]'./

\verb/entailment, neutral or contradiction? /

\verb/Output:/'',

where \texttt{[Statement1]}, \texttt{[Statement2]} and \texttt{[Answer]} are replaced by XVNLI test data. 

\subsection{Detailed results} \label{sec:detailed-res}

\Cref{tab:res-xflickrco,tab:res-xm3600,tab:res-xgqa,tab:res-maxm,tab:res-marvl,tab:res-xvnli,tab:res-marvl,tab:res-m30k,tab:res-commute} show the detailed results for all languages in which it can be observed that the model trained on mOSCAR outperforms the model trained on captions only by a large margin.

\begin{table*}[htbp]
    \centering
    \resizebox{.8\textwidth}{!}{\begin{tabular}{lccccccccc} \toprule
   & & \multirow{2}{*}{De} & \multirow{2}{*}{En} & \multirow{2}{*}{Es} & \multirow{2}{*}{Id} & \multirow{2}{*}{Ja} & \multirow{2}{*}{Ru} & \multirow{2}{*}{Tr} & \multirow{2}{*}{Zh} \\
  & \#shots  &  &  &  &  &  &  &  &  \\ 
    \midrule
         & 0  & 26.93 & 29.64 & 14.07 & 32.04 & 
 \hphantom{0}2.87 & 18.07 & \hphantom{0}4.23 & \hphantom{0}7.40 \\
      Multilingual OF   & 4  & 54.38 & 51.47 & 37.32 & 47.22 & 11.06 & 32.23 & 13.03 & 31.71 \\
      \textit{mOSCAR + caps.}   & 8  & 55.09 & 56.75 & 34.99 & \textbf{51.60} & 15.03 & 34.17 & 13.63 & 33.90 \\
       & 16  & \textbf{61.59} & \textbf{59.89} & \textbf{39.46} & 51.50 & \textbf{19.63} & \textbf{34.94} & \textbf{14.19} & \textbf{34.49} \\ \midrule

          & 0  & 16.72 & 24.57 & \hphantom{0}3.80 & 10.82 & \hphantom{0}2.82 & \hphantom{0}8.20 & \hphantom{0}2.79 & \hphantom{0}6.82 \\
          
      Multilingual OF   & 4  & 21.10 & 31.05 & \hphantom{0}7.52 & \hphantom{0}9.63 & \hphantom{0}3.84 & 13.21 & \hphantom{0}7.01 & 12.20 \\
      
      \textit{captions only}   & 8  & 32.56 & 35.73 & 13.35 & 15.85 & \hphantom{0}5.96 & 18.13 & \hphantom{0}6.97 & 15.47 \\
      
       & 16  & 29.86 & 40.57 & 13.75 & 23.83 &  \hphantom{0}6.92 & 20.40 & \hphantom{0}7.90 & 15.73 \\ \bottomrule

    \end{tabular}}
    \caption{Captioning results (CideR scores) on xFlickr\&CO. \textbf{Bold} is best result.}
    \label{tab:res-xflickrco}
\end{table*}

\begin{table*}[!htbp]
    \centering
    \resizebox{.8\linewidth}{!}{\begin{tabular}{lcccccccccccc}
        \toprule
   & & \multirow{2}{*}{Ar} & \multirow{2}{*}{Cs} & \multirow{2}{*}{Da} & \multirow{2}{*}{De} & \multirow{2}{*}{El} & \multirow{2}{*}{En} & \multirow{2}{*}{Es} & \multirow{2}{*}{Fa} & \multirow{2}{*}{Fi} & \multirow{2}{*}{Fr} & \multirow{2}{*}{He} \\
  & \#shots  &  &  &  &  &  &  &  & &  &  &  \\ 
    \midrule

   & 0  & \hphantom{0}4.83 & \hphantom{0}2.50 & \hphantom{0}8.52 & \hphantom{0}8.16 & \hphantom{0}0.76 & 42.57 & 16.79 & 12.49 & \hphantom{0}1.26 & 14.76 & \hphantom{0}3.76 \\
   
 Multi. OF  & 4  & 22.74 & \hphantom{0}6.42 & 33.73 & 24.29 & \hphantom{0}2.32 & 77.98 & 37.81 & 31.94 & \hphantom{0}6.78 & 39.79 & 15.51 \\
 
 \textit{full}  & 8  & 22.91 & \hphantom{0}7.41 & 35.23 &  \textbf{25.79} & \hphantom{0}2.95 & 77.64 & 38.41 & \textbf{35.46} & \hphantom{0}7.92 &  42.81 & 15.85 \\
 
   & 16  & \textbf{23.47} & \textbf{\hphantom{0}8.14} & \textbf{35.96} & \textbf{25.47} & \hphantom{0}2.58 & \textbf{78.18} & \textbf{39.18} & 31.44 & \textbf{\hphantom{0}8.42} & \textbf{43.77} & \textbf{16.08} \\ \midrule

   & 0  & \hphantom{0}2.24 & \hphantom{0}0.97 & \hphantom{0}6.42 & \hphantom{0}6.46 & \hphantom{0}3.68 & 10.02 & \hphantom{0}9.32 & \hphantom{0}4.95 & \hphantom{0}1.14 & 16.15 & \hphantom{0}0.78 \\
 Multi. OF  & 4  & \hphantom{0}5.36 & \hphantom{0}1.36 & 13.11 & 11.82 & \hphantom{0}7.78 & 35.52 & 19.96 & \hphantom{0}9.62 & \hphantom{0}1.86 & 22.48 & \hphantom{0}2.29  \\
 \textit{Caps only}  & 8  & \hphantom{0}6.76 & \hphantom{0}1.40 & 15.29 & 14.39 & \hphantom{0}7.21 & 37.28 & 21.90 & 12.19 & \hphantom{0}2.08 & 23.27 & \hphantom{0}1.71 \\
   & 16  & \hphantom{0}6.25 & \hphantom{0}2.29 & 17.96 & 15.11 & \textbf{\hphantom{0}7.64} &  48.03 & 25.39 & \hphantom{0}9.21 & \hphantom{0}2.10 & 30.16 & \hphantom{0}2.72 \\ \midrule \midrule

& & \multirow{2}{*}{Hi} & \multirow{2}{*}{Hr} & \multirow{2}{*}{Hu} & \multirow{2}{*}{Id} & \multirow{2}{*}{It} & \multirow{2}{*}{Ja} & \multirow{2}{*}{Ko} & \multirow{2}{*}{Nl} & \multirow{2}{*}{No} & \multirow{2}{*}{Pl} & \multirow{2}{*}{Pt} \\
  & \#shots  &  &  &  &  &  &  &  & &  &  &  \\ 
    \midrule

       & 0  & \hphantom{0}2.79 & \hphantom{0}2.00 & \hphantom{0}1.51 & \hphantom{0}9.96 & 11.53 & \hphantom{0}0.92 & \hphantom{0}0.58 & 16.11 & \hphantom{0}8.31 & \hphantom{0}3.94  & 13.37 \\
       
 Multi. OF  & 4  & 11.03 & 10.87 & \hphantom{0}5.87 & 25.88 & 29.53 & 17.45 & 10.85 & 46.22 & 25.18 & 15.36  & 31.32 \\
 
 \textit{full}  & 8  & 11.61 & \textbf{12.00} & \hphantom{0}6.91 & 29.68 & \textbf{29.34} & 20.13 & \textbf{12.01} & 47.58 & \textbf{27.08} & \textbf{17.80}  & \textbf{33.29} \\
 
   & 16  & \textbf{12.74} & 11.40 & \hphantom{0}7.03 & 26.73 & \textbf{30.43} & \textbf{20.57} & 11.07 & \textbf{49.33} & 27.07 & 17.15 & 32.79 \\ \midrule

   & 0  & \hphantom{0}2.29 & \hphantom{0}0.97 & \hphantom{0}3.51 & \hphantom{0}2.98 & \hphantom{0}7.96 & \hphantom{0}1.85 & \hphantom{0}1.05 & \hphantom{0}4.88 & \hphantom{0}5.78 & \hphantom{0}0.92 & \hphantom{0}9.79  \\
 Multi. OF  & 4  & \hphantom{0}4.57 & \hphantom{0}1.72 & \hphantom{0}7.57 & \hphantom{0}6.39 & 16.23 & \hphantom{0}3.47 & \hphantom{0}4.33 & 11.26 & 11.99 & \hphantom{0}1.16  & 15.93   \\
 \textit{Caps only}  & 8  & \hphantom{0}5.94 & \hphantom{0}2.17 & \hphantom{0}7.83 & \hphantom{0}9.93 &  15.40 & \hphantom{0}7.93 & \hphantom{0}5.34 & 11.87 & 13.79 & \hphantom{0}1.38 &  17.50  \\
   & 16  & \hphantom{0}6.36 & \hphantom{0}2.42 & \textbf{\hphantom{0}9.55} & 11.77 & 17.43 & 10.44 & \hphantom{0}6.03 & 12.98 & 14.65 & \hphantom{0}1.28 &  20.32  \\ \midrule \midrule

& & \multirow{2}{*}{Ro} & \multirow{2}{*}{Ru} & \multirow{2}{*}{Sv} & \multirow{2}{*}{Te} & \multirow{2}{*}{Th} & \multirow{2}{*}{Tr} & \multirow{2}{*}{Uk} & \multirow{2}{*}{Vi} & \multirow{2}{*}{Zh} &  &  \\
  & \#shots  &  &  &  &  &  &  &  & &  &  &  \\ 
    \midrule

       & 0  & \hphantom{0}1.84  & \hphantom{0}4.72 & 11.09 & \hphantom{0}0.88 & \hphantom{0}5.49 & \hphantom{0}2.86 & \hphantom{0}2.08 & 11.34 & \hphantom{0}3.29 &  &  \\
       
 Multi. OF  & 4  & \hphantom{0}6.08 & 21.46 & 30.24 & \hphantom{0}3.46 & 23.14 & 10.75 & 11.35 & 32.70 & 19.57 &  &  \\
 
 \textit{full}  & 8  & \textbf{\hphantom{0}7.10} & 21.78 & 30.26 &  \hphantom{0}3.76 & 25.17 & 12.83 & 12.26 & 35.86 & 20.11 &  &  \\
 
   & 16  & \hphantom{0}6.95 & \textbf{22.63} & \textbf{32.07} & \textbf{\hphantom{0}4.52} & \textbf{25.23} & \textbf{13.38} & \textbf{12.29} & \textbf{37.12} & \textbf{20.71} &  &  \\ \midrule

   & 0  & \hphantom{0}2.24 & \hphantom{0}1.93 & \hphantom{0}4.55 & \hphantom{0}0.67 & \hphantom{0}2.34 & \hphantom{0}2.68 & \hphantom{0}0.80 & \hphantom{0}8.55 & \hphantom{0}2.70 &  &  \\
 Multi. OF  & 4  & \hphantom{0}5.35 & \hphantom{0}6.29 & 15.66 & \hphantom{0}0.77 & \hphantom{0}7.21 & \hphantom{0}5.94 & \hphantom{0}1.76 & 20.69 & \hphantom{0}7.80 &  &  \\
 \textit{Caps only}  & 8  & \hphantom{0}5.18 & \hphantom{0}7.58 & 14.01 & \hphantom{0}1.00 & \hphantom{0}6.81 & \hphantom{0}8.90 & \hphantom{0}2.73 & 23.05 & \hphantom{0}8.99 &  &  \\
   & 16  & \hphantom{0}5.06 & \hphantom{0}9.06 & 20.60 & \hphantom{0}1.18 & \hphantom{0}8.35 &  10.25 & \hphantom{0}3.47 & 25.16 & 11.05 &  &  \\ \midrule \midrule

    \end{tabular}}
    \caption{Captioning results (CideR scores) on XM3600. \textbf{Bold} is best result.}
    \label{tab:res-xm3600}
\end{table*}

\begin{table*}[htbp]
    \centering
    \begin{tabular}{lccccccccc} \toprule
       & & \multirow{2}{*}{Bn} & \multirow{2}{*}{De} & \multirow{2}{*}{En} & \multirow{2}{*}{Id} & \multirow{2}{*}{Ko} & \multirow{2}{*}{Pt} & \multirow{2}{*}{Ru} & \multirow{2}{*}{Zh} \\
  & \#shots  &  &  &  &  &  &  &  &  \\ \midrule

                             & 0  & 22.76 & 25.72 & 34.24 & 26.68 & 26.89 & 26.73 & 25.28 & 27.32  \\
   Multilingual OF           & 4  & 26.72 & 32.57 & 37.91 & 32.54 & 31.88 & 32.35 & 31.28 & 33.4 \\
   \textit{mOSCAR + caps.}   & 8  & 28.07 & 35.15 & 39.44 & 35.14 & 32.94 & 35.59 & 33.58 & 34.04 \\
                             & 16  & \textbf{29.64} & \textbf{37.33} & \textbf{40.09} & \textbf{35.55} & \textbf{34.06} & \textbf{36.27} & \textbf{34.50} & \textbf{35.36}  \\ \midrule

                             & 0  & 10.54 & \hphantom{0}6.51 & 10.43 & \hphantom{0}7.74 & \hphantom{0}7.50 & \hphantom{0}7.79 & \hphantom{0}8.62 & \hphantom{0}9.84 \\
   Multilingual OF           & 4  & 12.54 & 11.90 & 15.78 & 13.95 & 13.70 & 12.01 & 12.73 & 15.03 \\
    \textit{captions only}   & 8  & 11.62 & 11.70 & 17.29 & 13.86 & 12.85 & 11.60 & 12.65 & 15.35 \\
                             & 16  & \hphantom{0}9.77 & 11.86 & 18.37 & 13.24 & 12.48 & 11.25 & 11.24 & 14.33 \\ \midrule \midrule
            \multicolumn{10}{c}{\textit{Translate Test}} \\ \midrule

    \multirow{4}{*}{OF-3B MPT}  & 0 & 18.64 & 18.67 & - & 18.36 & 17.54 & 19.21 & 18.88 & 17.11 \\
                                & 4 & 23.23 & 23.40 & - & 22.95 & 22.46 & 23.52 & 22.41 & 22.85 \\
                                & 8 & 28.22 & 29.44 & - & 28.21 & 27.67 & 29.58 & 28.21 & 28.63 \\
                                & 16 & 31.31 & 32.58 & - & 31.82 & 31.42 & 32.74 & 31.62 & 31.22 \\ \midrule

                                & 0 & 30.41 & 32.1 & - & 29.35 & 29.99 & 31.39 & 29.06 & 28.81 \\
       Multilingual OF          & 4 & 34.89 & 36.32 & - & 35.50 & 35.64 & 36.84 & 35.05 & 34.60 \\
       \textit{mOSCAR + caps.}  & 8 & 35.95 & 37.65 & - & 36.78 & 37.14 & 37.81 & 36.17 & 35.98 \\
                                & 16 & \textbf{36.78} & \textbf{38.78} & - &\textbf{37.52} & \textbf{37.73} & \textbf{38.68} & \textbf{37.91} & \textbf{36.84} \\ \bottomrule
   
    \end{tabular} \vspace{2mm}
    \caption{VQA results on xGQA. \textbf{Bold} is best result.}
    \label{tab:res-xgqa}
\end{table*}

\begin{table*}[htbp]
    \centering
    \begin{tabular}{lcccccccc} \toprule
        & & \multirow{2}{*}{En} & \multirow{2}{*}{Fr} & \multirow{2}{*}{Hi} & \multirow{2}{*}{He} & \multirow{2}{*}{Ro} & \multirow{2}{*}{Th} & \multirow{2}{*}{Zh} \\
  & \#shots  &  &  &  &  &  &  &    \\ \midrule

                            & 0 & 36.58 & 28.03 & 20.38 & 18.21 & 15.49 & 24.25 & 13.36 \\
   Multi. OF                & 4 & 38.13 & 30.03 & 23.08 & 21.43 & 17.61 & 31.72 & 22.02 \\
    \textit{mOSCAR + caps}  & 8 & 38.52 & 29.55 & 24.62 & 20.00 & 17.61 & \textbf{25.27} & \textbf{23.83} \\
                            & 16 & 35.80 & \textbf{31.82} & \textbf{25.00} & \textbf{23.93} & 19.01 & \textbf{33.96} & 22.74 \\ \midrule

                                & 0 & \hphantom{0}9.73 & \hphantom{0}0.38 & \hphantom{0}7.69 & \hphantom{0}1.43 & \hphantom{0}0.00 & \hphantom{0}5.22 & \hphantom{0}3.61 \\
      Multi. OF                 & 4 & \hphantom{0}9.34 & \hphantom{0}2.65 & \hphantom{0}5.00 & \hphantom{0}2.50 & \hphantom{0}0.00 & \hphantom{0}5.60 & \hphantom{0}3.97 \\
      \textit{captions only}    & 8 & \hphantom{0}9.34 & \hphantom{0}1.89 & \hphantom{0}8.08 & \hphantom{0}5.00 & \hphantom{0}1.06 & \hphantom{0}3.36 & \hphantom{0}5.42 \\
                                & 16 & \hphantom{0}8.56 & \hphantom{0}1.14 & \hphantom{0}5.00 & \hphantom{0}8.21 & \hphantom{0}0.35 & \hphantom{0}3.36 & \hphantom{0}7.58 \\ \midrule \midrule

    \multicolumn{9}{c}{\textit{Translate test}} \\ \midrule

       \multirow{4}{*}{OF-3B MPT}     & 0 & - & 12.50 & 22.31 & \hphantom{0}0.36 & 10.92 & \hphantom{0}0.00 & \hphantom{0}0.00 \\
                                      & 4 & - & 10.98 & 25.38 & \hphantom{0}0.36 & 10.21 & \hphantom{0}0.00 & \hphantom{0}0.00 \\
                                      & 8 & - & 10.98 & 27.31 & \hphantom{0}0.36 & 11.27 & \hphantom{0}0.00 & \hphantom{0}0.00 \\
                                      & 16 & - & 13.26 & 26.54 & \textbf{\hphantom{0}1.07} & 13.38 & \hphantom{0}0.00 & \hphantom{0}0.00 \\ \midrule

                                     & 0 & - & \textbf{18.18} & 28.08 & \hphantom{0}0.00 & 13.73 & \hphantom{0}0.00 & \textbf{\hphantom{0}0.36} \\
           Multi. OF                 & 4 & - & 15.91 & 30.38 & \hphantom{0}0.36 & 12.68 & \hphantom{0}0.00 & \hphantom{0}0.00 \\
            \textit{mOSCAR + caps}   & 8 & - & 15.15 & 30.77 & \hphantom{0}0.00 & 14.79 & \hphantom{0}0.00 & \hphantom{0}0.00 \\
                                     & 16 & - & 15.91 & \textbf{35.77} & \hphantom{0}0.36 & \textbf{16.90} & \hphantom{0}0.00 & \hphantom{0}0.00 \\ \bottomrule

    \end{tabular} \vspace{2mm}
    \caption{VQA results on MaXM. \textbf{Bold} is best result.}
    \label{tab:res-maxm}
\end{table*}

\begin{table*}[htbp]
\centering
    \begin{minipage}{.45\linewidth}
    \centering
    \resizebox{\textwidth}{!}{\begin{tabular}{lcccccc} \toprule
     & & \multirow{2}{*}{Id} & \multirow{2}{*}{Sw} & \multirow{2}{*}{Ta} & \multirow{2}{*}{Tr} & \multirow{2}{*}{Zh} \\
  & \#shots  &  &  &  &  &   \\ \midrule

    Random chance & & 50.00 & 50.00 & 50.00 & 50.00 & 50.00 \\ \midrule

                             & 0  & 50.09  & 49.46  & 49.60  & 49.83 & 48.81  \\ 
  Multilingual OF            & 4  & 49.91 & 48.19 & 49.68 & 50.42 &  50.00 \\ 
   \textit{mOSCAR + caps}    & 8  & \textbf{53.55} & \textbf{50.72} & 49.76 & \textbf{51.78} &  \textbf{51.58} \\ 
                             & 16  & 48.94 & 49.82 & 49.20 & 50.25 &  50.99 \\ \midrule

                             & 0  & 51.33 & 49.01 & 49.52 & 49.83 & 49.70  \\ 
   Multilingual OF           & 4  & 49.73 & 49.64 & 49.19 & 49.41 &  49.70 \\ 
   \textit{captions only}    & 8  & 49.91 & 49.10 & 49.60 & 49.75 &  49.90 \\ 
                             & 16  & 50.09 & 49.73 & 49.60 & 49.75 &  49.80 \\ \midrule \midrule

       \multicolumn{7}{c}{\textit{Translate test}} \\ \midrule

  \multirow{4}{*}{OF-3B MPT}   & 0  & 50.00 & 49.37 & 49.76 & 49.83 & 49.80  \\ 
                               & 4  & 50.00 & 49.64 & 49.52 & 49.75 &  49.60 \\ 
                               & 8  & 49.82 & 49.46 & 49.28 & 50.08 &  49.90 \\ 
                               & 16  & 50.00 & 49.37 & 49.44 & 50.00 &  49.80 \\ \midrule

                             & 0  & 49.07 & 49.79 & 49.52 & 50.34 &  49.60 \\ 
  Multilingual OF            & 4  & 49.99 & 49.79 & 48.23 & 49.75 &  49.76 \\ 
   \textit{mOSCAR + caps}    & 8  & 50.00 & 48.92 & \textbf{50.64} & 50.42 &  48.90 \\ 
                             & 16  & 49.84 & 50.00 & 50.24 & 48.90 &  49.75 \\ \bottomrule

    \end{tabular}}
    \caption{Classification results on MaRVL. \textbf{Bold} is best result.}
    \label{tab:res-marvl}
    \end{minipage}%
    \hspace{0.08\linewidth}
  \begin{minipage}{.45\linewidth}

    \resizebox{\textwidth}{!}{\begin{tabular}{lcccccc}\toprule
     & & \multirow{2}{*}{Ar} & \multirow{2}{*}{En} & \multirow{2}{*}{Es} & \multirow{2}{*}{Fr} & \multirow{2}{*}{Ru} \\
     
  & \#shots  &  &  &  &  &   \\ \midrule

    Random chance & & 33.33 & 33.33 & 33.33 & 33.33 & 33.33 \\ \midrule

                             & 0 & 33.51 & 34.62 & 33.08 & 34.02 & 34.19 \\
  Multilingual OF.           & 4 & 33.08 & 33.59 & 33.42 & 34.45 & 35.82 \\
   \textit{mOSCAR + caps.}   & 8 & \textbf{35.91} & \textbf{38.75} & \textbf{35.14} & \textbf{36.08} & \textbf{37.11} \\
                             & 16 & 34.11 & 36.60 & 33.93 & 34.54 & 35.05 \\ \midrule

                             & 0 & 35.48 & 34.02 & 33.51 & 34.45 & 31.36 \\
   Multilingual OF.          & 4 & 32.04 & 31.79 & 32.73 & 32.22 & 31.44 \\
   \textit{captions only}    & 8 & 34.02 & 33.76 & 32.04 & 35.57 & 33.16 \\
                             & 16 & 32.04 & 32.99 & 33.76 & 33.17 & 31.53 \\ \midrule \midrule

    \multicolumn{7}{c}{\textit{Translate test}} \\ \midrule

   \multirow{4}{*}{OF-3B MPT}  & 0 & 32.65 & - & 31.01 & 31.44 & 35.82 \\
                               & 4 & 36.25 & - & 35.82 & 35.57 & 35.65 \\
                               & 8 & 31.27 & - & 31.10 & 31.10 & 31.70 \\
                               & 16 & 33.68 & - & 33.25 & 32.99 & 33.25 \\ \midrule

                             & 0 & 34.88 & - & 34.88 & 34.54 & 34.36 \\
  Multilingual OF.           & 4 & 36.25 & - & 36.17 & 35.91 & 36.08 \\
   \textit{mOSCAR + caps.}   & 8 & \textbf{39.60} & - & \textbf{39.52} & \textbf{40.29} & \textbf{39.35} \\
                             & 16 & 37.54 & - & 37.89 & 37.46 & 39.00 \\ \bottomrule
    
    \end{tabular}}
    \caption{Classification results on XVNLI. \textbf{Bold} is best result.}
    \label{tab:res-xvnli}
    \end{minipage}
\end{table*}

\begin{table*}[hbtp]
    \centering
    \begin{minipage}{.45\linewidth}
    \centering
    \begin{tabular}{lcccc} \toprule
        & & \multirow{2}{*}{Cs} & \multirow{2}{*}{De} & \multirow{2}{*}{Fr} \\
        & \#shots  &  &  &   \\ \midrule

                        & 0 & 2.82 & 28.45 & 37.47 \\
     Multi. OF          & 4 & 3.12 & 29.20 & 37.49 \\
      \textit{full}     & 8 & 3.14 & \textbf{29.62} & 37.99 \\
                        & 16 & \textbf{3.34} & 29.41 & \textbf{38.79} \\ \midrule

                            & 0 & 0.00 & \hphantom{0}0.00 & \hphantom{0}0.00 \\
     Multi. OF              & 4 & 0.00 & \hphantom{0}0.00 & \hphantom{0}0.00 \\
      \textit{caps. only}   & 8 & 0.00 & \hphantom{0}0.00 & \hphantom{0}0.03 \\
                            & 16 & 0.00 & \hphantom{0}0.40 & \hphantom{0}1.82 \\ \bottomrule
   
    \end{tabular} \vspace{2mm}
    \caption{En$\rightarrow$X translation results on Multi30k. \textbf{Bold} is best result.}
    \label{tab:res-m30k}
    \end{minipage}%
    \hspace{0.08\linewidth}
  \begin{minipage}{.45\linewidth}
    \begin{tabular}{lcccc} \toprule
        & & \multirow{2}{*}{Cs} & \multirow{2}{*}{De} & \multirow{2}{*}{Fr} \\
        & \#shots  &  &  &   \\ \midrule

                        & 0 & 56.49 & \textbf{65.67} & 67.86 \\
     Multi. OF          & 4 & 57.47 & 64.00 & \textbf{68.18} \\
      \textit{full}     & 8 & 58.44 & 64.33 & 67.86 \\
                        & 16 & 58.11 & 62.67 & 66.23 \\ \midrule

                            & 0 & 58.12 & 61.67 & 64.29 \\
     Multi. OF              & 4 & \textbf{59.09} & 61.00 &  63.31 \\
      \textit{caps. only}   & 8 & \textbf{59.09} & 59.34 & 64.29 \\
                            & 16 & 58.12 & 58.67 & 63.96 \\ \bottomrule
   
    \end{tabular} \vspace{2mm}
    \caption{En$\rightarrow$X CoMMuTE results. \textbf{Bold} is best result.}
    \label{tab:res-commute}
    \end{minipage}
\end{table*}

\FloatBarrier

\clearpage

\begin{table*}[!ht]
\centering
\resizebox{\linewidth}{!}{\begin{tabular}{lccccccccc}
\toprule
& \# shots & xFlickR\&CO & XM3600 & xGQA & MaXM & MaRVL & XVNLI & Multi30k & CoMMuTE \\ \midrule
\multirow{2}{*}{InternVL2 4B} & 0 & 16.21 & \hphantom{0}7.02 & 12.38 & \hphantom{0}6.35 & 53.14 & 33.85 & 26.99 & 66.93 \\ 
& 4 & 24.89 & \hphantom{0}9.53 & 26.05 & 14.72 & 54.22 & 35.72 & 26.68 & 64.22 \\ \midrule
PaliGemma 3B & 0 & 28.28 & \textbf{24.49} & \textbf{42.68} & \textbf{33.42} & 51.48 & 39.36 & 17.98 & 62.78 \\ \midrule
Idefics2 8B & 0 & 27.11 & 15.94 & 22.53 & 28.99 & \textbf{63.18} & \textbf{50.33} & \textbf{30.19} & \textbf{67.13} \\ \midrule
Llava-NeXT 8B & 0 & 23.67 & 14.70 & 25.48 & 15.17 & 60.50 & 45.40 & 29.40 & 66.37 \\ \midrule \midrule
\multirow{2}{*}{Multi. OF 3B (\textit{ours})} & 0 & 16.91 & \hphantom{0}7.45 & 26.95 & 22.23 & 49.56 & 33.88 & 22.91 & 63.34 \\
& 4 & \textbf{34.80} & 22.18 & 32.23 & 26.33 & 49.64 & 34.07 & 23.27 & 63.22 \\ \bottomrule
\end{tabular}}
\caption{Results averaged across languages. \textbf{Bold} is best result.}
\label{tab:res-sota}
\end{table*}

\subsection{Comparison with state-of-the-art mLLMs}\label{sec:res-sota}

We computed the results for different state-of-the-art models of similar sizes as multilingual Open Flamingo namely: (1) InternVL2-4B\footnote{\texttt{OpenGVLab/InternVL2-4B}} (2) PaliGemma\footnote{\texttt{google/paligemma-3b-pt-224}} (3) Idefics2-8B\footnote{\texttt{HuggingFaceM4/idefics2-8b}} and (3) Llava-NeXT 8B\footnote{\texttt{llava-hf/llama3-llava-next-8b-hf}}. InternVL2 and PaliGemma are trained on multilingual and multimodal data while Llava-NeXT and Idefics2 are trained on English multimodal datasets.

Table~\ref{tab:res-sota} shows results averaged across languages for different state-of-the-art mLLMs of sizes from 3b to 8B. These results highlights multiple things: (1) getting results significantly better than random (MaRVL and XVNLI) requires instruction-tuning data as Idefics2 and Llava-NeXT were both trained on instruction-tuning multimodal datasets (2) English-only still gets decent results on multilingual benchmarks despite not having been trained on multilingual and multimodal data, probably due to their underlying LLM being multilingual (3) multilingual Open Flamingo (trained on mOSCAR and captions) gets superior results to InternVL2-4B on VQA benchmarks and captioning benchmarks but inferior to PaliGemma-3B mainly due to the fact that it was trained on much less data and the quality of the captions used to train multilingual Open Flamingo may not be as good as the WebLI dataset used to train PaliGemma.


\end{document}